\documentclass{article} 

\pdfoutput=1

\usepackage{iclr2023_conference,times}

\usepackage{mathtools} 
\usepackage{booktabs} 
\usepackage{tikz} 
\usepackage{amsmath}
\usepackage{amsfonts}
\usepackage{multirow}
\usepackage[font=small,labelfont=bf]{caption}
\usepackage{color,colortbl}
\usepackage{xcolor}
\usepackage{xspace}
\usepackage{subfigure}
\usepackage{bbm}
\usepackage{bm}

\usepackage{hyperref}
\usepackage{url}

\newcommand{\algoNameFull}{SlotFormer\xspace}

\definecolor{lightgrey}{rgb}{0.43,0.43,0.43}
\definecolor{crimson}{rgb}{0.86,0.08,0.24}
\hypersetup{
    colorlinks,
    citecolor=lightgrey,
    linkcolor=crimson,
    urlcolor=lightgrey
}

\definecolor{LavenderBlue}{rgb}{0.7020,    0.8039,    0.8902}
\definecolor{Lightapricot}{rgb}{0.9961,    0.8510,    0.6510}
\definecolor{thirdtablecolor}{rgb}{0.8706,    0.7961,    0.8941}

\newcommand{\heading}[1]{\noindent\textbf{#1}}

\long\def\ignorethis#1{}

\definecolor{demphcolor}{RGB}{100,100,100}
\newcommand{\demph}[1]{\textcolor{demphcolor}{#1}}
\newcommand{\pms}[2]{#1{\tiny{{\demph{{$\pm$#2}}}}}}

\newlength\paramargin
\newlength\figmargin
\newlength\tablemargin
\newlength\secmargin
\newlength\figcapmargin
\newlength\rowmargin
\newlength\tablecapmargin
\newlength\pagetopmargin

\usepackage{titlesec}
\titlespacing{\section}{0pt}{0.3\baselineskip}{0.2\baselineskip}
\titlespacing{\subsection}{0pt}{0.15\baselineskip}{0.1\baselineskip}
\titlespacing{\subsubsection}{0pt}{0.05\baselineskip}{0.03\baselineskip}

\renewcommand{\paragraph}[1]{\vspace{0.2em}\noindent\textit{#1} --}

\usepackage{soul}

\newcommand{\revised}[1]{#1}

\title{\algoNameFull: Unsupervised Visual Dynamics \\ Simulation with Object-Centric Models}

\author{
Ziyi Wu$^{1,2}$, Nikita Dvornik$^{3,1}$, Klaus Greff$^{4}$, Thomas Kipf\thanks{Equal advisory contribution} $^{4}$, Animesh Garg\footnotemark[1] $^{1,2}$ \\
$^1$ University of Toronto, $^2$ Vector Institute, $^3$ Samsung AI Centre Toronto, $^4$ Google Research\\
}

\iclrfinalcopy
\begin{document}

\maketitle

\vspace{-4mm}  
\begin{abstract}

Understanding dynamics from visual observations is a challenging problem that requires disentangling individual objects from the scene and learning their interactions.
While recent object-centric models can successfully decompose a scene into objects, modeling their dynamics effectively still remains a challenge.
We address this problem by introducing \algoNameFull \ -- a Transformer-based autoregressive model operating on learned object-centric representations.
Given a video clip, our approach reasons over object features to model spatio-temporal relationships and predicts accurate future object states.
In this paper, we successfully apply \algoNameFull to perform video prediction on datasets with complex object interactions.
Moreover, the unsupervised \algoNameFull's dynamics model can be used to improve the performance on supervised downstream tasks, such as Visual Question Answering (VQA), and goal-conditioned planning.
Compared to past works on dynamics modeling, our method achieves significantly better long-term synthesis of object dynamics, while retaining high quality visual generation.
Besides, \algoNameFull enables VQA models to reason about the future without object-level labels, even outperforming counterparts that use ground-truth annotations.
Finally, we show its ability to serve as a world model for model-based planning, which is competitive with methods designed specifically for such tasks.
Additional results and details are available at our \href{https://slotformer.github.io/}{Website}.

\end{abstract}

\section{Introduction}

The ability to understand complex systems and interactions between its elements is a key component of intelligent systems.
Learning the dynamics of a multi-object systems from visual observations entails capturing \textit{object} instances, their appearance, position and motion, and simulating their spatio-temporal interactions.
Both in robotics~\citep{CDNA,SAVP} and computer vision~\citep{ConvLSTM,PredRNN}, unsupervised learning of dynamics has been a central problem due to its important practical implications.
Obtaining a faithful dynamics model of the environment enables future prediction, planning and, crucially, allows to transfer the dynamics knowledge to improve downstream supervised tasks, such as visual reasoning~\citep{DCL,VRDP}, planning~\citep{sun2022plate} and model-based control~\citep{TransformerWorldModel}.
Yet, an effective domain-independent approach for unsupervised visual dynamics learning from video remains elusive.

One approach to visual dynamics modeling is to frame it as a prediction problem directly in the pixel space~\citep{ConvLSTM,PredRNN,SVG-LP}.
This paradigm builds on global frame-level representations, and uses dense feature maps of past frames to predict future features.
By design, such models are object-agnostic, treating background and foreground modeling as equal.
This frequently results in poorly learned object dynamics, producing unrealistic future predictions over longer horizons~\citep{VPReview}.
Another perspective to dynamics learning is through object-centric dynamics models~\citep{SQAIR,R-NEM,STOVE}.
This class of methods first represents a scene as a set of object-centric features (a.k.a.~slots), and then learns the interactions among the slots to model scene dynamics.
It allows for more natural dynamics modeling and leads to more faithful simulation~\citep{OP3,PARTS}.
To achieve this goal, earlier object-centric models bake in strong scene~\citep{SCALOR} or object~\citep{G-SWM} priors in their frameworks, while more recent methods~\citep{CSWM,PARTS} learn object interactions purely from data, with the aid of Graph Neural Networks (GNNs)~\citep{GNNSurvey} or Transformers~\citep{Attention}.
Yet, these approaches independently model the per-frame object interactions and their temporal evolution, using different networks. 
This suggests that a simpler and more effective dynamics model is yet to be designed.

In this work, we argue that learning a system's dynamics from video effectively requires two key components: i) \textit{strong unsupervised object-centric representations} (to capture objects in each frame) and ii) a \textit{powerful dynamical module} (to simulate spatio-temporal interactions between the objects).
To this end, we propose \algoNameFull: an elegant and effective Transformer-based object-centric dynamics model, which builds upon object-centric features~\citep{SAVi,STEVE}, and requires no human supervision.
We treat dynamics modeling as a sequential learning problem: given a sequence of input images, \algoNameFull takes in the object-centric representations extracted from these frames, and predicts the object features in the future steps.
By conditioning on multiple frames, our method is capable of capturing the spatio-temporal object relationships simultaneously, thus ensuring consistency of object properties and motion in the synthesized frames.
We evaluate \algoNameFull on four video datasets consisting of diverse object dynamics.
Our method not only presents competitive results on standard video prediction metrics, but also achieves significant gains when evaluating on object-aware metrics in the long range.
Crucially, we demonstrate that \algoNameFull's unsupervised dynamics knowledge can be successfully transferred to downstream supervised tasks (e.g., VQA and goal-conditional planning) to improve their performance ``for free".
In summary, this work makes the following contributions:
\begin{enumerate}
    \item  \algoNameFull: a Transformer-based model for object-centric visual simulation;
    \item  \algoNameFull achieves state-of-the-art performance on two video prediction datasets, with significant advantage in modeling long-term dynamics;
    \item \algoNameFull achieves state-of-the-art results on two VQA datasets and competitive results in one planning task, when equipped with a corresponding task-specific readout module. 
\end{enumerate}

\section{Related Work}

In this section, we provide a brief overview of related works on physical reasoning, object-centric models and Transformers, which is further expanded in Appendix~\ref{appendix:related-work}.

\heading{Dynamics modeling and intuitive physics.}
Video prediction methods treat dynamics modeling as an image translation problem~\citep{ConvLSTM,PredRNN,SVG-LP,SAVP}, and model changes in the pixel space.
However, methods that model dynamics using global image-level features usually struggle with long-horizon predictions.
Some approaches leverage local priors~\citep{CDNA,SNA}, or extra input information~\citep{flowvp1,posevp1}, which only help in the short term.
More recent works improve modeling visual dynamics using explicit object-centric representations.
Several works directly learn deep models in the abstracted state space of objects~\citep{Galileo,InteractionNetwork,BilliardsVP,NeuralPhysicsEngine}.
However, they require ground-truth physical properties for training, which is unrealistic for visual dynamics simulation.
Instead, recent works use object features from a supervised detector as the base representation for visual simulation~\citep{CVP,PropNet,RPIN,yu2022mac} with a GNN-based dynamics model.
In contrast to the above works, our model is completely unsupervised; \algoNameFull belongs to the class of models that learn both object discovery and scene dynamics without supervision.
We review this class of models below.

\heading{Unsupervised object-centric representation learning from videos.}
Our work builds upon recent efforts in decomposing raw videos into temporally aligned \textit{slots}~\citep{SAVi,SIMONe,STEVE}.
Earlier works often make strong assumptions on the underlying object representations.
\citet{SCALOR} explicitly decompose the scene into foreground and background to apply fixed object size and presence priors.
\citet{G-SWM} further disentangle object features to represent object positions, depth and semantic attributes separately.
Some methods leverage the power of GNNs or Transformers to eliminate these domain-specific priors~\citep{OP3,R-NEM,OAT,PARTS}. However, they still model the object interactions and temporal scene dynamics using separate modules; and set the context window of the recurrent dynamics module to only a single timestep.
The most relevant work to ours is OCVT~\citep{OCVT}, which also applies Transformers to slots from multiple frames.
However, OCVT utilizes manually disentangled object features, and needs Hungarian matching for latent alignment during training.
Therefore, it still underperforms RNN-based baselines in the video prediction task.
In contrast, \algoNameFull is a general Transformer-based dynamics model which is agnostic to the underlying object-centric representations.
It performs joint spatio-temporal reasoning over object slots simultaneously, enabling consistent long-term dynamics modeling.

\heading{Transformers for sequential modeling.}
Inspired by the success of autoregressive Transformers in language modeling~\citep{GPT-1,GPT-2,GPT-3}, they were adapted to video generation tasks~\citep{VideoGPT,lookoutside,TransformerWorldModel,Transframer}.
To handle the high dimensionality of images, these methods often adopt a two-stage training strategy by first mapping images to discrete tokens~\citep{VQGAN}, and then learning a Transformer over tokens.
However, since they operate on a regular image grid, the mapping ignores the boundary of objects and usually splits one object into multiple tokens.
In this work, we learn a Transformer-based dynamics model over \textit{slot}-based representations that capture the entire object in a single vector, thus generating more consistent future object states as will be shown in the experiments.

\section{\algoNameFull: Object-Oriented Dynamics Learning}

\begin{figure*}[t]
    \vspace{\pagetopmargin}
    \vspace{-1.5mm}
    \centering
    \includegraphics[width=0.95\linewidth]{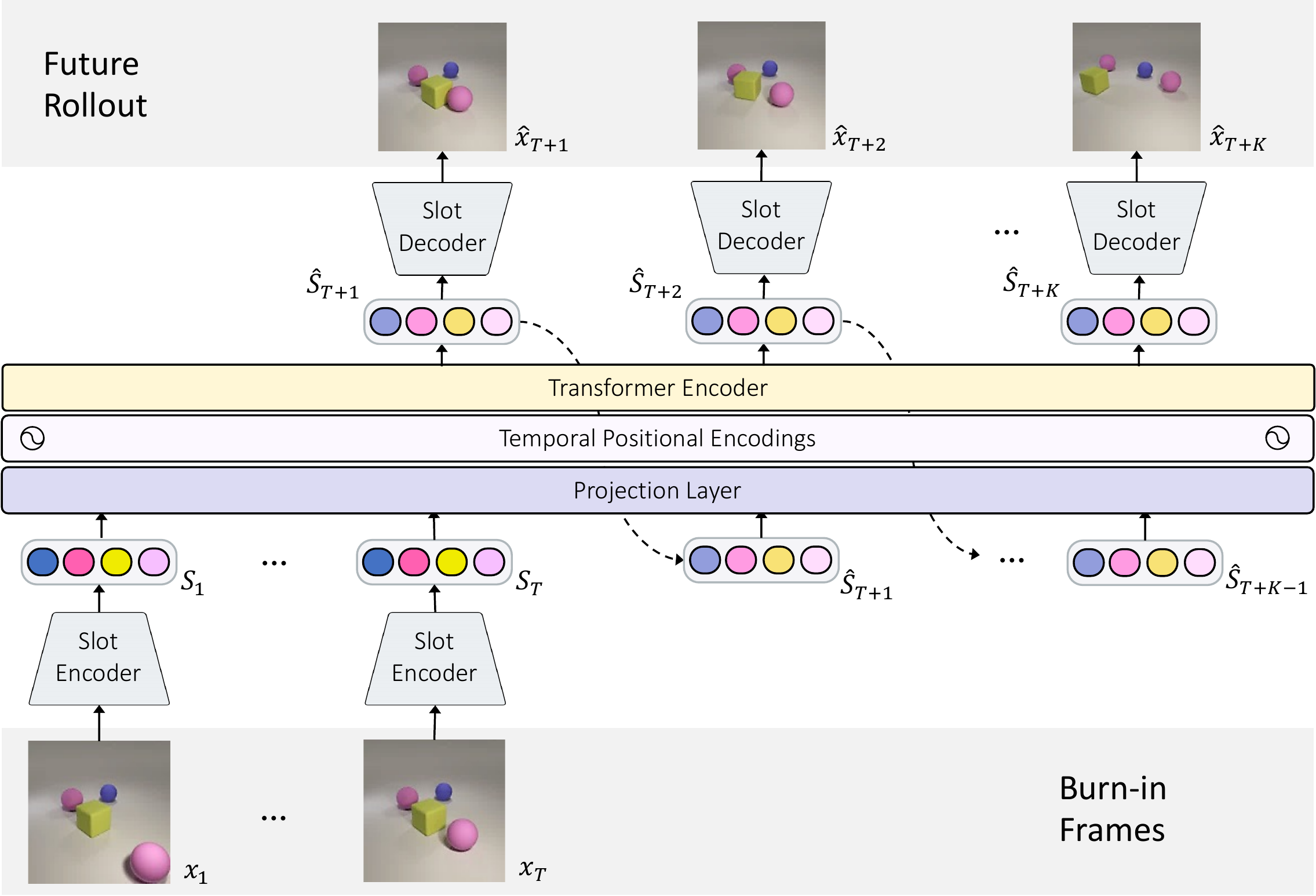}
    \caption{
    \algoNameFull architecture overview.
    Taking multiple video frames $\{\bm{x}_t\}_{t=1}^T$ as input, we first extract object slots $\{\mathcal{S}_t\}_{t=1}^T$ using the pretrained object-centric model.
    Then, slots are linearly projected and added with temporal positional encoding.
    The resulting tokens are fed to the Transformer module to generate future slots $\{\hat{\mathcal{S}}_{T+k}\}_{k=1}^K$ in an autoregressive manner.
    }
    \label{fig:model-pipeline}
    \vspace{-1.5mm}
\end{figure*}

Taking $T$ video frames as inputs, \algoNameFull first leverages a pre-trained object-centric model to extract object slots from each frame (Section~\ref{sec:savi}).
These slots are then forwarded to the Transformer module for joint spatio-temporal reasoning, and used to predict future slots autoregressively (Section~\ref{sec:transformer}).
The whole pipeline is trained by minimizing the reconstruction loss in both feature and image space (Section~\ref{sec:loss}).
The overall model architecture is illustrated in Figure~\ref{fig:model-pipeline}.

\subsection{Slot-based Object-Centric Representation}\label{sec:savi}

We build on the Slot Attention architecture to extract slots from videos due to their strong performance in unsupervised object discovery.
Given $T$ input frames $\{\bm{x}_t\}_{t=1}^T$, our object-centric model first extracts image features using a Convolutional Neural Network (CNN) encoder, then adds positional encodings, and flattens them into a set of vectors $\bm{h}_t \in \mathbb{R}^{M \times D_{enc}}$, where $M$ is the size of the flattened feature grid and $D_{enc}$ is the feature dimension.
Then, the model initializes $N$ slots $\tilde{\mathcal{S}}_t \in \mathbb{R}^{N \times D_{slot}}$ from a set of learnable vectors ($t = 1$), and performs Slot Attention~\citep{SlotAttn} to update the slot representations as $\mathcal{S}_t = f_{SA}(\tilde{\mathcal{S}}_t, \bm{h}_t)$.
Here, $f_{SA}$ binds slots to objects via iterative Scaled Dot-Product Attention~\citep{Attention}, encouraging scene decomposition.
To achieve temporal alignment of slots, $\tilde{\mathcal{S}}_t$ for $t \ge 2$ is initialized as $\tilde{\mathcal{S}}_{t} = f_{trans}(\mathcal{S}_{t-1})$, where $f_{trans}$ is the transition function implemented as a Transformer encoder.

Before training the Transformer-based dynamics model, we first pre-train the object-centric model using reconstruction loss on videos from the target dataset.
This ensures the learned slots can accurately capture both foreground objects and background environment of the scene.

\subsection{Dynamics Prediction with Autoregressive Transformer}\label{sec:transformer}

\heading{Overview.}
Given slots $\{\mathcal{S}_t\}_{t=1}^T$ extracted from $T$ video frames, \algoNameFull is able to synthesize a sequence of future slots $\{\mathcal{S}_{T+k}\}_{k=1}^K$ for any given horizon $K$.
Our model operates by alternating between two steps: i) feed the slots into a Transformer that performs joint spatio-temporal reasoning and predicts slots at the next timestep, $\hat{\mathcal{S}}_{t+1}$,
ii) feed the predicted slots back into the Transformer to keep generating future rollout autoregressively.
See Figure~\ref{fig:model-pipeline} for the pipeline overview.

\heading{Architecture.}
To build the \algoNameFull's dynamics module, $\mathcal{T}$, we adopt the standard Transformer encoder module with $N_\mathcal{T}$ layers.
To match the inner dimensionality $D_e$ of $\mathcal{T}$, we linearly project the input sequence of slots to a latent space $G_t = \mathrm{Linear}(\mathcal{S}_t) \in \mathbb{R}^{N \times D_e}$.
To indicate the order of input slots, we add positional encoding (P.E.) to the latent embeddings.
A naive solution would be to add a sinusoidal positional encoding to every slot regardless of its timestep, as done in \citet{Aloe}.
However, this would break the \textit{permutation equivariance} among slots, which is a useful property of our model.
Therefore, we only apply positional encoding at the temporal level, such that the slots at the same timestep receives the same positional encoding:
\begin{equation}
    V = [G_1, G_2, ..., G_T] + [P_1, P_2, ..., P_T],
\end{equation}
where $V \in \mathbb{R}^{(TN) \times D_e}$ is the resulting input to the transformer $\mathcal{T}$ and $P_t \in \mathbb{R}^{N \times D_e}$ denotes the sinusoidal positional encoding duplicated $N$ times.
As we will show in the ablation study, the temporal positional encoding enables better prediction results despite having fewer parameters.

Now, we can utilize the Transformer $\mathcal{T}$ to reason about the dynamics of the scene.
Denote the Transformer output features as $U = [U_1, U_2, ..., U_T] \in \mathbb{R}^{(TN) \times D_e}$, we take the last $N$ features $U_T \in \mathbb{R}^{N \times D_e}$ and feed them to a linear layer to obtain the predicted slots at timestep $T + 1$:
\begin{equation}
    U = \mathcal{T}(V),\ \ \ \ \hat{\mathcal{S}}_{T+1} = \mathrm{Linear}(U_{T}).
\end{equation}
For consequent future predictions, $\hat{\mathcal{S}}_{T+1}$ will be treated as the ground-truth slots along with $\{\mathcal{S}_t\}_{t=2}^T$ to predict $\hat{\mathcal{S}}_{T+2}$.
In this way, the Transformer can be applied autoregressively to generate any given number, $K$, of future frames, as illustrated in Figure~\ref{fig:model-pipeline}.

\textit{Remark.}
The \algoNameFull's architecture allows to \textit{preserve temporal consistency among slots} at different timesteps.
To realize such consistency, we employ residual connections from $\mathcal{S}_t$ to $\hat{\mathcal{S}}_{t+1}$, which forces the Transformer $\mathcal{T}$ to apply refinement to the slots while preserving their absolute order.
Owing to this order invariance, \algoNameFull can be used to reason about individual object's dynamics for long-term rollout, and can be seamlessly integrated with downstream task models.

\subsection{Model Training}\label{sec:loss}

\revised{
Error accumulation is a key issue in long-term generation~\citep{VPReview}.
In contrast to prior works~\citep{OCVT} that use a GPT-style causal attention mask~\citep{GPT-1}, \algoNameFull predicts all the slots at one timestep in parallel.
Therefore, we train the model using the predicted slots as inputs, simulating the autoregressive generation process at test time.
This reduces the train-test discrepancy, thus improving the quality of long-term generation as shown in Section~\ref{sec:exp-ablation}.
}

For training, we use a slot reconstruction loss (in $L_2$) denoted as:
\begin{equation}
    \mathcal{L}_S = \frac{1}{K \cdot N} \sum_{k=1}^K \sum_{n=1}^N ||\hat{\bm{s}}_{T+k}^n - \bm{s}_{T+k}^n||^2.
\end{equation}
When using SAVi as the object-centric model, we also employ an image reconstruction loss to promote prediction of consistent object attributes such as colors and shapes.
The predicted slots are decoded to images by the frozen SAVi decoder $f_{dec}$, and then matched to the original frames as:
\begin{equation}
    \mathcal{L}_I = \frac{1}{K} \sum_{k=1}^K ||f_{dec}(\hat{\mathcal{S}}_{T+k}) - \bm{x}_{T+k}||^2.
\end{equation}
The final objective function is a weighted combination of the two losses with a hyper-parameter $\lambda$:
\begin{equation}
    \mathcal{L} = \mathcal{L}_S + \lambda \mathcal{L}_I.
\end{equation}

\section{Experiments}

\algoNameFull is a generic architecture for many tasks requiring object-oriented reasoning. 
We evaluate the dynamics modeling capability of \algoNameFull in three such tasks: video prediction, VQA and action planning.
Our experiments aim to answer the following questions:
    \textbf{(1)} Can an autoregressive Transformer operating on slots generate future frames with both high visual quality and accurate object dynamics? (Section~\ref{sec:video-prediction})
    \textbf{(2)} Are the future states synthesized by \algoNameFull useful for reasoning in   VQA? (Section~\ref{sec:vqa})
    \textbf{(3)} How well can \algoNameFull serve as a world model for planning actions? (Section~\ref{sec:planning})
Finally, we perform an ablation study of \algoNameFull's components in Section~\ref{sec:exp-ablation}.

\subsection{Experimental Setup}

\heading{Datasets.}
We evaluate our method's capability in video prediction on two datasets, \textit{OBJ3D}~\citep{G-SWM} and \textit{CLEVRER}~\citep{CLEVRER}, and demonstrate its ability for downstream reasoning and planning tasks on three datasets, \textit{CLEVRER}, \textit{Physion}~\citep{Physion} and \textit{PHYRE}~\citep{PHYRE}.
We briefly introduce each dataset below, which are further detailed in Appendix~\ref{appendix:dataset}.

\textit{OBJ3D} consists of CLEVR-like~\citep{CLEVR} dynamic scenes, where a sphere is launched from the front of the scene to collide with other still objects.
There are 2,920 videos for training and 200 videos for testing.
Following~\citep{G-SWM}, we use the first 50 out of 100 frames of each video in our experiments, since most of the interactions end before 50 steps.

\textit{CLEVRER} is similar to OBJ3D but with smaller objects and varying entry points throughout the video, making it more challenging.
For video prediction evaluation, we follow \citet{PARTS} to subsample the video by a factor of 2, resulting in a length of 64.
We also filter out video clips where there are newly entered objects during the rollout period.
For VQA task, CLEVRER provides four types of questions: descriptive, explanatory, predictive and counterfactual.
The predictive questions require the model to simulate future interactions of objects such as collisions.
Therefore, we focus on the accuracy improvement on predictive questions by using \algoNameFull's future rollout.

\textit{Physion} is a VQA dataset containing realistic simulation of eight physical phenomena.
Notably, Physion features diverse object entities and environments, making physical reasoning more difficult than previous synthetic VQA benchmarks.
The goal of this dataset is to predict whether a red \textit{agent} object will contact with a yellow \textit{patient} object when the scene evolves.
Following the official evaluation protocol, all models are first trained using unsupervised future prediction loss, then used to perform rollout on test scenes, where a linear readout model is applied to predict the answer.

\textit{PHYRE} is a physical reasoning benchmark consisting of 2D physical puzzles.
We use the BALL-tier, where the goal is to place a red ball at a certain location, such that the green ball will eventually come in contact with the blue/purple object, after the scene is unrolled in time.
Following \citet{RPIN}, we treat \algoNameFull as the world model and build a task success classifier on predicted object states as the scoring function.
Then, we use it to rank a pre-defined 10,000 actions from \citet{PHYRE}, and execute them accordingly.
We experiment on the within-template setting.

\heading{Implementation Details.}
We first pre-train the object-centric model on each dataset, and then extract slots for training \algoNameFull.
We employ SAVi~\citep{SAVi} on OBJ3D, CLEVRER, PHYRE, and STEVE~\citep{STEVE} on Physion to extract object-centric features.
\revised{
SAVi leverages a CNN decoder to reconstruct videos from slots, while STEVE uses a Transformer-based slot decoder.
STEVE can perform scene decomposition on visually complex data, but requires more memory and training time.
}
Besides, we discovered that vanilla SAVi cannot handle some videos on CLEVRER.
So we also introduce a stochastic version of SAVi to solve this problem. 
\revised{
Please refer to Appendix~\ref{appendix:implement-ours} and their papers for complete details of network architectures and hyper-parameters.
}

\subsection{Evaluation on Video Prediction}\label{sec:video-prediction}
\revised{
In this subsection, we evaluate \algoNameFull's ability in long-term visual dynamics simulation.
We train all models on short clips cropped from videos, and rollout for longer steps during evaluation.
}

\begin{figure*}[t]
    \vspace{\pagetopmargin}
    \vspace{-5mm}
    \centering
    \subfigure{
        \includegraphics[width=0.48\textwidth]{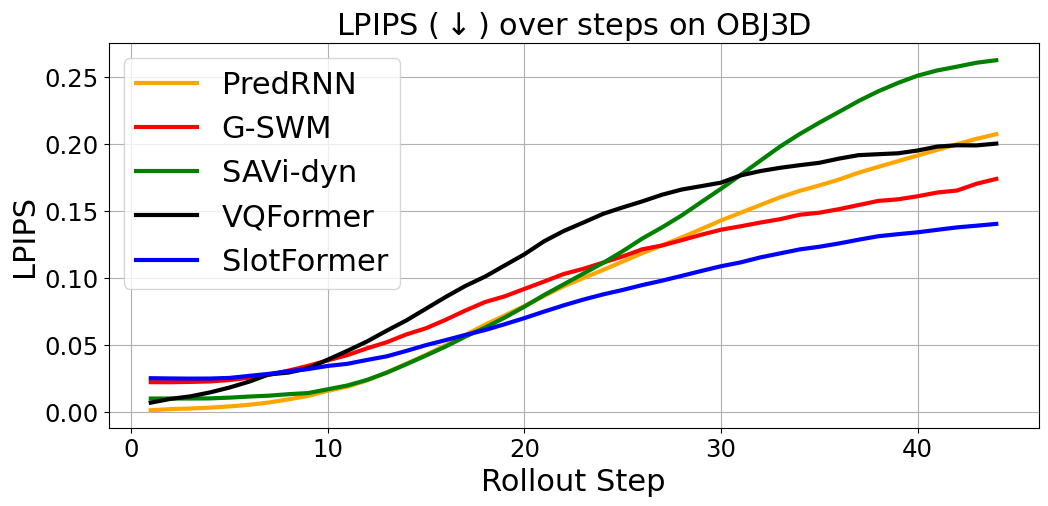}
    }
    \subfigure{
        \includegraphics[width=0.48\textwidth]{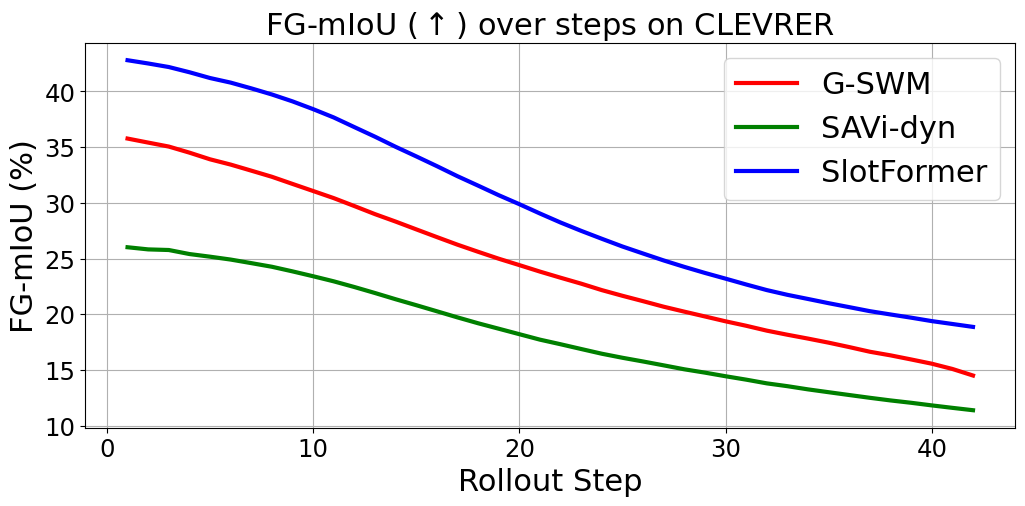}
    }
    \vspace{-2.5mm}
    \caption{
    Video dynamics modeling with \algoNameFull as a function of future steps.
    (left) Visual quality of decoded frames measured with LPIPS and (right) the quality of decoded foreground object masks with mIoU.
    }
    \label{fig:per-step-results}
    \vspace{\figmargin}
    \vspace{-0.5mm}
\end{figure*}

\begin{table}[t]
    \begin{minipage}{.55\linewidth}
    \small
    \resizebox{0.99\textwidth}{!}{
    \setlength{\tabcolsep}{2pt}
    \begin{tabular}{lccc|ccc}
        \toprule
        \multirow{2}{*}{\textbf{Method}} & \multicolumn{3}{c}{\textbf{OBJ3D}} & \multicolumn{3}{c}{\textbf{CLEVRER}} \\
        & PSNR $\uparrow$ & SSIM $\uparrow$ & LPIPS $\downarrow$ & PSNR $\uparrow$ & SSIM $\uparrow$ & LPIPS $\downarrow$ \\
        \midrule
        PredRNN & \textbf{33.68} & 0.91 & 0.12 & \textbf{31.34} & \textbf{0.90} & 0.17 \\
        SAVi-dyn & 32.94 & 0.91 & 0.12 & 29.77 & 0.89 & 0.19 \\
        G-SWM & 31.43 & 0.89 & 0.10 & 28.42 & 0.89 & 0.16 \\
        VQFormer & 30.71 & 0.86 & 0.11 & 26.80 & 0.85 & 0.18 \\
        \midrule
        \textbf{Ours} & 32.40 & 0.91 & \textbf{0.08} & 30.21 & 0.89 & \textbf{0.11} \\
        \bottomrule
    \end{tabular}
    }
    \caption{
    Evaluation of visual quality on both datasets.
    \label{table:vp-visual-quality}
    }
    \end{minipage}%
    \hfill
    \begin{minipage}{.445\linewidth}
    \small
    \resizebox{0.99\textwidth}{!}{
    \setlength{\tabcolsep}{2pt}
    \begin{tabular}{lcccc}
        \toprule
        \textbf{Method} & AR $\uparrow$ & ARI $\uparrow$ & FG-ARI $\uparrow$ & FG-mIoU $\uparrow$ \\
        \midrule
        SAVi-dyn & 8.94 & 8.64 & \textbf{64.32} & 18.25 \\
        G-SWM & 43.98 & 57.14 & 49.61 & 24.44 \\
        \midrule
        \textbf{Ours} & \textbf{53.14} & \textbf{63.45} & 63.00 & \textbf{29.81} \\
        \bottomrule
    \end{tabular}
    }
    \caption{
    Evaluation of object dynamics on CLEVRER.
    All the numbers are in $\%$.
    \label{table:vp-object-dynamics}
    }
    \end{minipage}
    \vspace{\tablemargin}
    \vspace{-0.5mm}
\end{table}

\begin{figure*}[!t]
    \centering
    \includegraphics[width=0.95\linewidth]{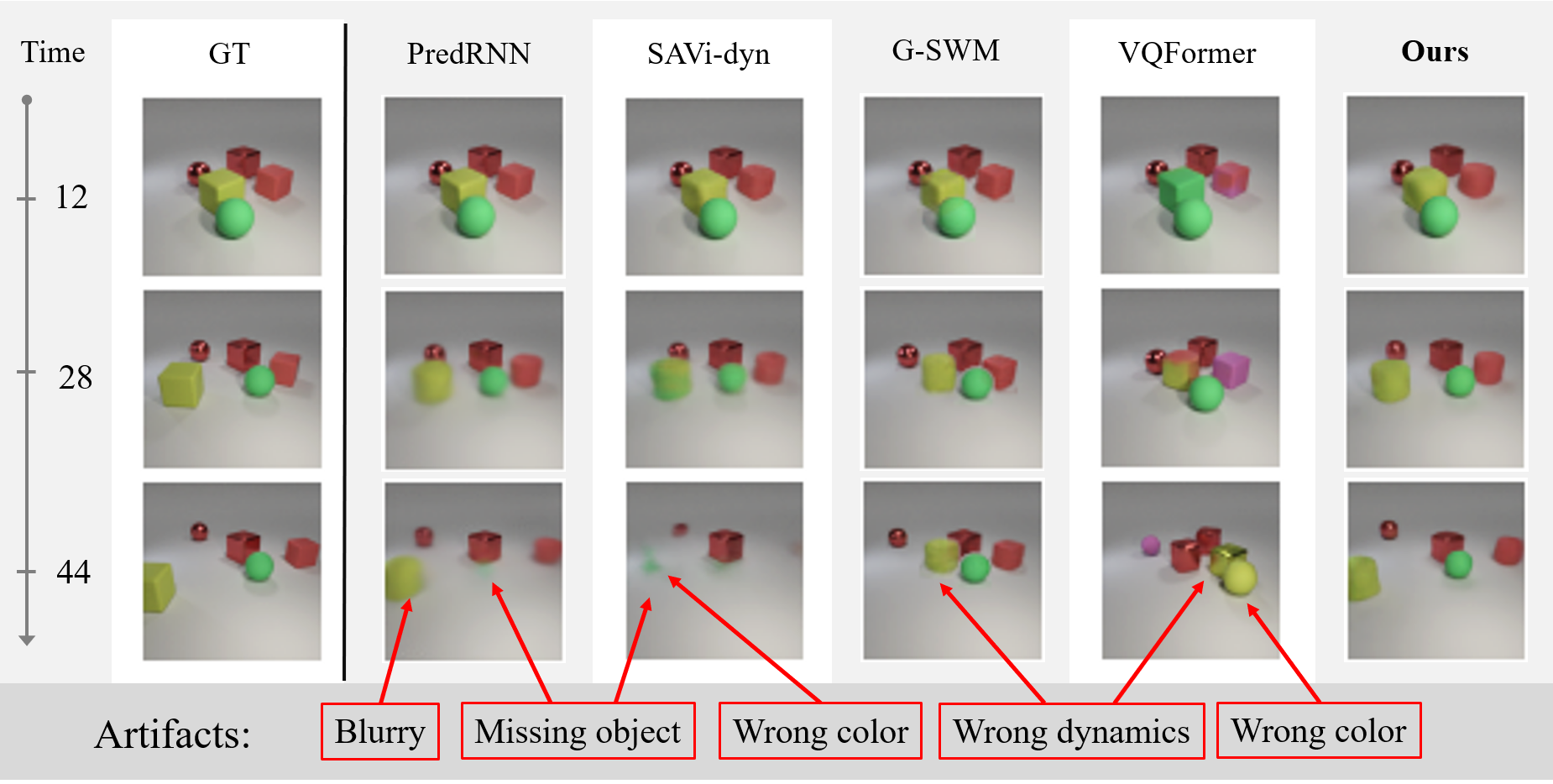}
    \caption{
    Generation results on OBJ3D.
    Despite higher PSNR, PredRNN and SAVi-dyn produce images with artifacts, while \algoNameFull simulates sharp frames and accurate object dynamics.
    }
    \label{fig:obj3d-vis}
    \vspace{-1mm}
\end{figure*}

\heading{Baselines.}
We compare our approach with four baselines which are further described in Appendix~\ref{appendix:baselines}.
We use a video prediction model \textit{PredRNN}~\citep{PredRNN} that generates future frames based on global image features.
To verify the effectiveness of slot representation, we train a VQ-VAE~\citep{VQVAE} to tokenize images, and replace the slot in \algoNameFull with patch tokens, denoted as \textit{VQ-Former}.
We also adopt the state-of-the-art generative object-centric model \textit{G-SWM}~\citep{G-SWM}, which applies heavy priors.
Finally, since the PARTS~\citep{PARTS} code is unreleased, we incorporate their Transformer-LSTM dynamics module into SAVi (denoted as \textit{SAVi-dyn}) and train the model with the same setup.
\revised{
We include additional baselines~\citep{OCVT} in Appendix~\ref{appendix:more-baseline}.
}

\heading{Evaluation Metrics.}
To evaluate the visual quality of the videos, we report PSNR, SSIM~\citep{SSIM} and LPIPS~\citep{LPIPS}.
As discussed in \citet{ImageQuality}, LPIPS captures better perceptual similarity with human than PSNR and SSIM.
We focus our comparison on LPIPS, while reporting others for completeness.
It is worth noting that neither of these metrics evaluate semantics in predicted frames~\citep{yu2022mac}. 
To evaluate the predicted object dynamics, we use the per-slot object masks predicted by the SAVi decoder and compare them to the ground-truth segmentation mask; same for the corresponding bounding box.
We calculate the Average Recall (AR) of the predicted object boxes, and the Adjusted Rand Index (ARI), the foreground variant of ARI and mIoU termed FG-ARI and FG-mIoU of the predicted masks.
We unroll the model for 44 and 42 steps on OBJ3D and CLEVRER, respectively, and report metrics averaged over timesteps.

\heading{Results on visual quality.}
Table~\ref{table:vp-visual-quality} presents the results on visual quality of the generated videos.
\algoNameFull outperforms all baselines with a sizeable margin in terms of LPIPS, and achieves competitive results on PSNR and SSIM.
We note that PSNR and SSIM are poor metrics in this setting.
For example, PredRNN and SAVi-dyn score highly in these two metrics despite producing blurry objects (see Figure~\ref{fig:obj3d-vis}).
In contrast, \algoNameFull generates objects with consistent attributes throughout the rollout, which we attribute to modeling dynamics in the object-centric space, rather than in the frames directly.
This is also verified in the per-step LPIPS results in Figure~\ref{fig:per-step-results} (left).
Since \algoNameFull relies on pretrained slots, the reconstructed images at earlier steps have lower quality than baselines.
Nevertheless, it achieves clear advantage at longer horizon, demonstrating superior long-term modeling ability.
Although VQFormer is also able to generate sharp images, it fails to predict correct dynamics and object attributes, as also observed in previous works~\citep{VideoGPT,lookoutside}. This shows that \textit{only a strong decoder (i.e. VQ-VAE) to generate realistic images is not sufficient} for learning dynamics.
See Appendix~\ref{appendix:more-vis} for more qualitative results on both datasets.

\heading{Results on object dynamics.}
Here, we evaluate the quality of object bounding boxes and segmentation masks, decoded from the models' future predictions.
The accuracy of the predicted object boxes and segmentation masks is summarized in Table~\ref{table:vp-object-dynamics} (right).
Since OBJ3D lacks such annotations, and PredRNN, VQFormer cannot generate object-level outputs, we exclude it from evaluation.
\algoNameFull achieves the best performance on AR, ARI and FG-mIoU, and competitive results on FG-ARI.
SAVi-dyn scores a high FG-ARI because its blurry predictions assign many background pixels to foreground objects, while the computation of FG-ARI ignores false positives.
This is verified by its poor performance in FG-mIoU which penalizes such mistakes.
We also show the per-step results in Figure~\ref{fig:per-step-results} (right) and Appendix~\ref{appendix:per-step-dynamics}, where our method excels at all future timesteps.

\heading{Attention map analysis.}
To study how \algoNameFull leverages past information to predict the future, we visualize the self-attention maps from the Transformer $\mathcal{T}$, which is detailed in Appendix~\ref{appendix:exp-attention-map}.

\begin{table}[t]
    \vspace{\pagetopmargin}
    \vspace{-2.5mm}
    \begin{minipage}{.48\linewidth}
        \centering
        \small
        \setlength{\tabcolsep}{3pt}
        \begin{tabular}{lcc}
        \toprule
        \textbf{Method} & per opt. ($\%$) & per ques. ($\%$) \\
        \midrule
        DCL & 90.5 & 82.0 \\
        VRDP & 91.7 & 83.8 \\
        VRDP$^\dag$ & 94.5 & 89.2 \\
        Aloe$^*$ & 93.1 & 87.3 \\
        \midrule
        Aloe$^*$ + \textbf{Ours} & \textbf{96.5} & \textbf{93.3} \\
        \bottomrule
    \end{tabular}
    \caption{
    Predictive VQA on CLEVRER, reporting per-option (per opt.) and per-question (per ques.) accuracy.
    DCL and VRDP$^\dag$ both utilize pre-trained object detectors;
    $^*$ indicates our re-implementation.
    \label{table:clevrer-vqa-numbers}
    }
    \end{minipage}%
    \hfill
    \begin{minipage}{.48\linewidth}
        \centering
        \small
        \setlength{\tabcolsep}{4pt}
        \begin{tabular}{lcc|c}
        \toprule
        \textbf{Method} & Obs. ($\%$) & Dyn. ($\%$) & $\uparrow$ ($\%$) \\
        \midrule
        \textcolor{gray}{Human} & \textcolor{gray}{74.7} & \textcolor{gray}{-} & \textcolor{gray}{-} \\
        \midrule
        RPIN$^*$ & 62.8 & 63.8 & +1.0 \\
        pDEIT-lstm$^*$ & 59.2 & 60.0 & +0.8 \\
        \midrule
        \textbf{Ours} & \textbf{65.2} & \textbf{67.1} & \textbf{+1.9} \\
        \bottomrule
    \end{tabular}
    \caption{
    VQA accuracy on Physion.
    We report the readout accuracy on observation (OBS.) and observation plus rollout (Dyn.) frames.
    $\uparrow$ denotes the improvement brought by the learned dynamics.
    Methods marked with $^*$ are our reproduced results.
    \label{table:physion-vqa-numbers}
    }
    \end{minipage}
    \vspace{\tablemargin}
\end{table}

\begin{figure*}[t]
    \vspace{-1mm}
    \centering
    \includegraphics[width=0.95\linewidth]{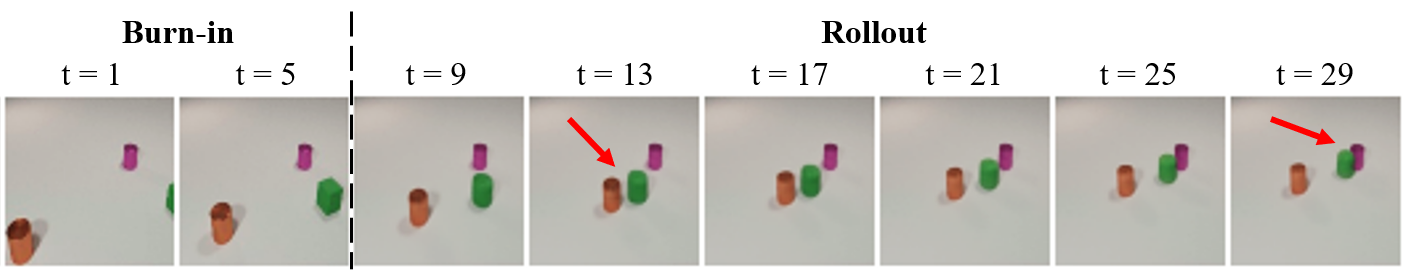}
    \caption{
    Qualitative results on CLEVRER VQA task.
    To answer the question ``\textit{Will the green object collide with the purple cylinder?}", \algoNameFull successfully simulates the first collision between the green and the brown cylinder (t = 13), which leads to the second collision between the target objects (t = 29).
    }
    \label{fig:vqa-vis}
    \vspace{-1mm}
\end{figure*}

\subsection{Visual Question Answering}\label{sec:vqa}
In this subsection, we show how to leverage (unsupervised) \algoNameFull's future predictions to improve (supervised) predictive question answering.

\heading{Our Implementation.}
On CLEVRER, we choose \textit{Aloe}~\citep{Aloe} as the base reasoning model as it can jointly process slots and texts.
\revised{
Aloe runs Transformers over slots from input frames and text tokens of the question, followed by an MLP to predict the answer.
}
For predictive questions, we explicitly unroll \algoNameFull and run Aloe on the predicted future slots.
See Appendix~\ref{appendix:aloe} for more details.
On Physion, since there is no language involved, we follow the official protocol by training a linear readout model on synthesized slots to predict whether the two objects contact.
We design an improved readout model for object-centric representations, which is further detailed in Appendix~\ref{appendix:physion-readout}.

\heading{Baselines.}
On CLEVRER, we adopt \textit{DCL}~\citep{DCL} which utilizes pre-trained object detectors and a GNN-based dynamics model.
We also choose the state-of-the-art model \textit{VRDP}~\citep{VRDP}, which exploits strong environmental priors to run differentiable physics engine for rollout.
We report two variants of VRDP which use Slot Attention (VRDP) or pre-trained detectors (VRDP$^\dag$) to detect objects.
Finally, for consistency with our results, we report the performance of our re-implemented Aloe (dubbed as Aloe$^*$).
\\
On Physion, we select \textit{RPIN}~\citep{RPIN} and \textit{pDEIT-lstm}~\citep{DEIT}, since they are the only two methods where the rollout improves accuracy in the benchmark~\citep{Physion}.
RPIN is an object-centric dynamics model using ground-truth bounding boxes.
pDEIT-lstm builds LSTM over ImageNet~\citep{ImageNet} pre-trained DeiT model, learning the dynamics over frame features.
Since the benchmark code for Physion is not released, we reproduce it to achieve similar or better results.
We also report the \textit{Human} results from the Physion paper for reference.

\heading{Results on CLEVRER.}
Table~\ref{table:clevrer-vqa-numbers} presents the accuracy on predictive questions.
The dynamics predicted by \algoNameFull boosts the performance of Aloe by 3.4\% and 6.0\% in the per option (per opt.) and per question (per ques.) setting, respectively.
As a fully unsupervised dynamics model, our method even outperforms previous state-of-the-art DCL and VRDP which use supervisedly trained object detectors.
On the CLEVRER public leaderboard
predictive question subset, we rank first in the per option setting, and second in the per question setting.
Figure~\ref{fig:vqa-vis} shows an example of our predicted dynamics, where \algoNameFull accurately simulates two consecutive collision events.

\heading{Results on Physion.}
Table~\ref{table:physion-vqa-numbers} summarizes the readout accuracy on observation (Obs.) and observation plus rollout (Dyn.) frames.
\algoNameFull achieves a 1.9\% improvement with learned dynamics, surpassing all the baselines.
See Figure~\ref{appendix-fig:vqa-vis} in the Appendix for qualitative results.

\begin{figure*}[t]
    \vspace{\pagetopmargin}
    \vspace{-5mm}
    \centering
    \subfigure[Slot decomposition on the first frame]{
        \includegraphics[width=0.38\textwidth]{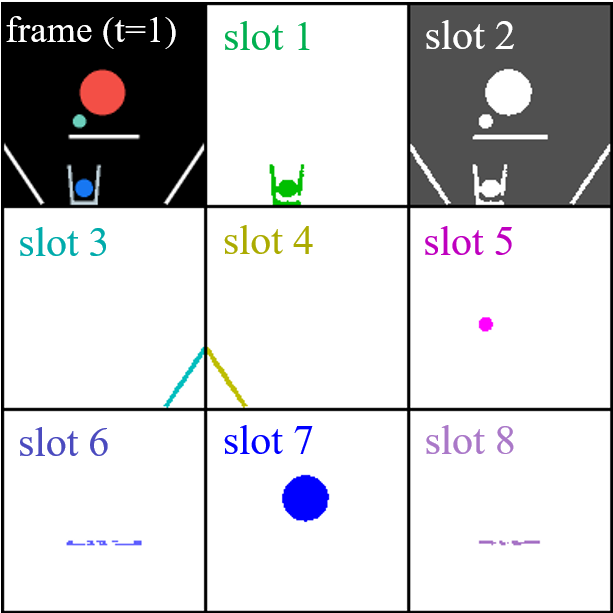}
    }
    \subfigure[Rollout results. Per-slot future predictions are color-coded.]{
        \includegraphics[width=0.6\textwidth]{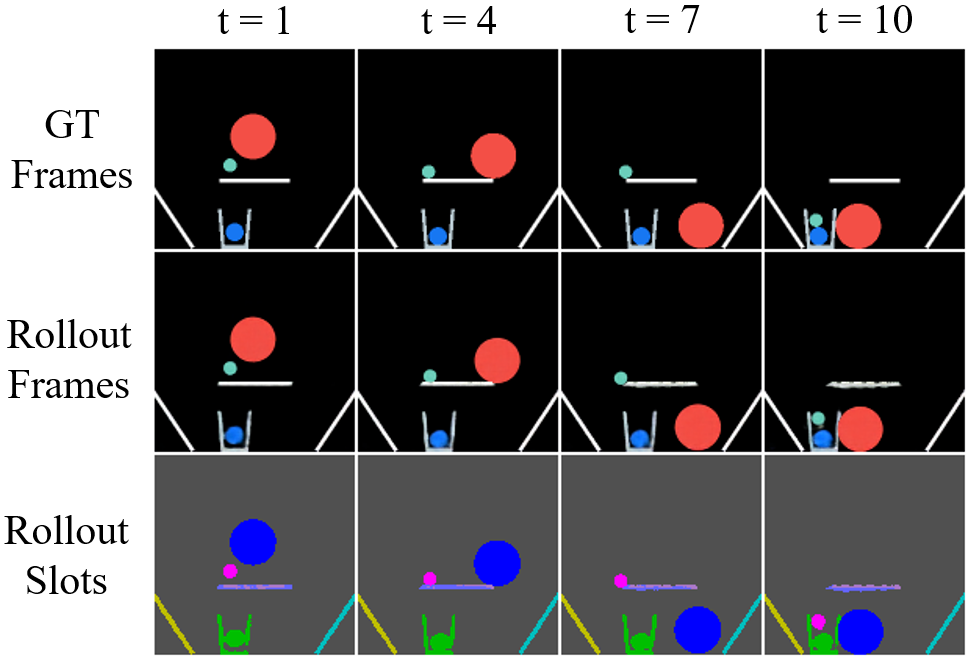}
    }
    \caption{
    Qualitative results on PHYRE.
    The goal is to place a red ball in the first frame, so that the green ball hits the blue object after rollout.
    We show the per-slot rollout, where \algoNameFull is able to decompose the scene into individual objects, and reason their interactions to perform accurate future synthesis.
    }
    \label{fig:phyre-vis}
    \vspace{\figmargin}
\end{figure*}

\begin{table}[t]
    \centering
    \small
    \setlength{\tabcolsep}{5pt}
    \begin{tabular}{ccccccc|cc}
        \toprule
        \textbf{Method} & RAND & MEM & DQN & Dec [Joint] & RPIN & Dyn-DQN & \revised{SAVi} & \textbf{Ours} \\
        \midrule
        \textbf{Annotations} & - & - & Mask & Mask & Mask & Mask & \revised{-} & - \\
        \midrule
        \textbf{AUCCESS} & \pms{13.7}{0.5} & \pms{2.4}{0.3} & \pms{77.6}{1.1} & \pms{80.0}{1.2} & \pms{85.2}{0.7} & \textbf{\pms{86.2}{0.9}} & \revised{\pms{80.7}{1.0}} & \pms{82.0}{1.1} \\
        \bottomrule
    \end{tabular}
    \caption{
    AUCCESS on PHYRE-1B within-template setting.
    All baseline learning methods use ground-truth object segmentation masks, while \algoNameFull is the only unsupervised technique learning from raw images.
    \label{table:phyre-numbers}
    }
\end{table}

\subsection{Action Planning}\label{sec:planning}
Here, we perform goal-conditioned planning inside the \algoNameFull's learned dynamics model.

\heading{Our Implementation.}
Since it is possible to infer the future states from only the initial configuration on PHYRE, we set the burn-in length $T$ = $1$, and apply \algoNameFull to generate slots $\mathcal{S}_2$ from $\mathcal{S}_1$.
Then, instead of only using $\mathcal{S}_2$ to generate $\mathcal{S}_3$, we feed in both $\mathcal{S}_1$ and $\mathcal{S}_2$ for better temporal consistency.
We apply this iterative overlapping modeling technique~\citep{lookoutside}, and set the maximum conditioning length as 6.
To rank the actions during testing, we train a task success classifier on future states simulated by \algoNameFull, which is detailed in Appendix~\ref{appendix:phyre-readout}.
We experiment on the within-template setting, and report the \textit{AUCCESS} metric averaged over the official 10 folds.
\revised{
AUCCESS measures the Area Under Curve (AUC) of the task success rate vs number of attempts curve for the first 100 attempted actions.
Please refer to Appendix~\ref{appendix-phyre-dataset} for more details.
}

\heading{Baselines.}
We report three naive baselines from \citet{PHYRE}, \textit{RAND}, \textit{MEM} and \textit{DQN}.
We adopt \textit{Dec [Joint]}~\citep{PHYREDecBaseline} which employs a CNN-based future prediction model, and \textit{RPIN}~\citep{RPIN} as an object-centric dynamics model.
Finally, Dynamics-Aware DQN~\citep{PHYREDynAware} (dubbed \textit{Dyn-DQN}) designs a task-specific loss to utilize dynamics information.
Notably, all of the above methods use either ground-truth object masks or bounding boxes, while \algoNameFull learns scene dynamics without any object-level annotations.
\revised{
To show the performance gain by rollout, we also report \textit{SAVi} which predicts task success purely from the initial frame's slots.
}

\heading{Results on action planning.}
We report the AUCCESS metric in Table~\ref{table:phyre-numbers}.
As an unsupervised dynamics model, \algoNameFull achieves an AUCCESS score of 82.0, which \revised{improves the non-rollout counterpart SAVi by 1.3, and} is on par with baselines that assume ground-truth object information as input.
Figure~\ref{fig:phyre-vis} shows the entire rollout generated by our model.
\algoNameFull is able to capture objects with varying appearance, and simulate the dynamics of complex multi-object interactions.

\begin{table}[t]
    \vspace{\pagetopmargin}
    \begin{minipage}{.51\linewidth}
        \centering
        \small
        \setlength{\tabcolsep}{3pt}
        \begin{tabular}{lccc}
        \toprule
        \textbf{Method} & PSNR $\uparrow$ & SSIM $\uparrow$ & LPIPS $\downarrow$ \\
        \midrule
        Ours (Full Model) & \textbf{32.40} & \textbf{0.91} & \textbf{0.080} \\
        \quad\quad Burn-in $T = 3$ & 31.26 & 0.88 & 0.093 \\
        \quad\quad Burn-in $T = 4$ & 31.95 & 0.89 & 0.088 \\
        \quad\quad Burn-in $T = 8$ & 32.08 & 0.90 & 0.082 \\
        \quad\quad Trans. Layer $N_\mathcal{T} = 8$ & 32.12 & 0.89 & 0.087 \\
        \quad\quad Naive P.E. & 32.05 & 0.90 & 0.082 \\
        \quad\quad Teacher Forcing & 30.52 & 0.87 & 0.106 \\
        \quad\quad No $\mathcal{L}_I$ & 31.23 & 0.88 & 0.093 \\
        \bottomrule
    \end{tabular}
    \caption{
    Ablation study on OBJ3D.
    \label{table:obj3d-ablation}
    }
    \end{minipage}%
    \hfill
    \begin{minipage}{.45\linewidth}
        \centering
        \small
        \setlength{\tabcolsep}{1pt}
        \begin{tabular}{lc}
        \toprule
        \textbf{Method} & Improved Acc. ($\%$) \\
        \midrule
        Ours (Full Model) & \textbf{1.9} \\
        \quad\quad Burn-in $T = 10$ & 1.0 \\
        \quad\quad Rollout $K = 5$ & 0.5 \\
        \quad\quad Rollout $K = 15$ & 1.9 \\
        \quad\quad Trans. Layer $N_\mathcal{T} = 4$ & 1.3 \\
        \quad\quad Trans. Layer $N_\mathcal{T} = 12$ & Diverge \\
        \quad\quad Naive P.E. & 1.6 \\
        \quad\quad Teacher Forcing & 0.2 \\
        \bottomrule
    \end{tabular}
    \caption{
    Ablation study on Physion.
    \label{table:physion-ablation}
    }
    \end{minipage}
\end{table}

\subsection{Ablation Study}\label{sec:exp-ablation}

In this section, we perform an ablation study to examine the importance of each component in \algoNameFull on OBJ3D (Table~\ref{table:obj3d-ablation}) and Physion (Table~\ref{table:physion-ablation}).

\heading{Burn-in length $T$ and rollout length $K$.}
By default, we set $T$ = $6$, $K$ = $10$ for OBJ3D, and $T$ = $15$, $K$ = $10$ for Physion.
On OBJ3D, the model performance first improves with more input frames, and slightly drops when $T$ further increases to $8$.
This might because a history length of 6 is sufficient for modeling accurate dynamics on OBJ3D.
On Physion, the accuracy improves consistently as we increase $T$, until using all the observation frames.
Besides, using 10 rollout frames strikes a balance between accuracy and computation efficiency.
\revised{
See Appendix~\ref{appendix:ablation} for line plots of these results.
}

\heading{Transformer (Trans.) Layer $N_\mathcal{T}$.}
By default, we set $N_\mathcal{T}$ = $4$ on OBJ3D and $N_\mathcal{T}$ = $8$ on Physion.
Stacking more layers harms the performance on OBJ3D due to overfitting, while improving the accuracy on Physion.
This is because the dynamics on Physion is more challenging to learn.
However, further increasing $N_\mathcal{T}$ to 12 makes model training unstable and the loss diverging.

\heading{Positional encoding (P.E.).}
Using a vanilla sinusoidal positional encoding which destroys the permutation equivariance among slots results in small performance drop in terms of visual quality, and a clear degradation in VQA accuracy.
This is not surprising, as permutation equivariance is a useful prior for object-centric scene modeling, which should be preserved.

\heading{Teacher forcing.}
We try the teacher forcing strategy~\citep{GPT-1} by taking in ground-truth slots instead of the predicted slots autoregressively during training, which degrades the results significantly.
This proves that simulating error accumulation benefits long-term dynamics modeling.

\heading{Image reconstruction loss $\mathcal{L}_I$.}
As shown in the table, the auxiliary image reconstruction loss improves the quality of the generated videos drastically.
As we observe empirically, $\mathcal{L}_I$ helps preserve object attributes (e.g., color, shape), but has little effect on object dynamics.
Thus, we do not apply $\mathcal{L}_I$ on Physion, due to the large memory consumption of STEVE's slot decoder.

\section{Conclusion}

In this paper, we propose \algoNameFull, a Transformer-based autoregressive model that enables consistent long-term dynamics modeling with object-centric representations.
\algoNameFull learns complex spatio-temporal interactions between the objects and generates future predictions of high visual quality.
Moreover, \algoNameFull can transfer unsupervised dynamics knowledge to downstream (supervised) reasoning tasks which leads to state-of-the-art or comparable results on VQA and goal-conditioned planning.
Finally, we believe that unsupervised object-centric dynamics models hold great potential for simulating complex datasets, advancing world models, and reasoning about the future with minimal supervision; and that \algoNameFull is a new step towards this goal.
\revised{
We discuss the limitations and potential future directions of this work in Appendix~\ref{appendix:limitations}.
}

\clearpage

\section*{Reproducibility Statement}

All of our methods are implemented in PyTorch~\citep{PyTorch}, and can be trained on servers with 4 modern GPUs in less than 5 days, enabling both industrial and academic researchers.
To ensure the reproducibility of our work, we provide detailed descriptions of how we process each dataset in Appendix~\ref{appendix:dataset}, the implementation details and hyper-parameters of the models we use in Appendix~\ref{appendix:implement-ours}, and sources of the baselines we compare with in Appendix~\ref{appendix:baselines}.
To facilitate future research, we will release the code of our work \revised{and the pre-trained model weights} alongside the camera ready version of this paper.

\subsubsection*{Acknowledgments}

We would like to thank Wei Yu, Tianyu Hua for general advice and feedback on the paper, Xiaoshi Wu, Weize Chen for discussion of Transformer model implementation and training, and Jiaqi Xi, Ritviks Singh, Quincy Yu, Calvin Yu, Liquan Wang for valuable discussions and support in computing resources.
Animesh Garg is supported by CIFAR AI chair, NSERC Discovery Award, University of Toronto XSeed Grant and NSERC Exploration grant. We would like to acknowledge Vector institute for computation support.

\bibliography{references}

\begin{thebibliography}{81}
\providecommand{\natexlab}[1]{#1}
\providecommand{\url}[1]{\texttt{#1}}
\expandafter\ifx\csname urlstyle\endcsname\relax
  \providecommand{\doi}[1]{doi: #1}\else
  \providecommand{\doi}{doi: \begingroup \urlstyle{rm}\Url}\fi

\bibitem[Abadi et~al.(2016)Abadi, Barham, Chen, Chen, Davis, Dean, Devin,
  Ghemawat, Irving, Isard, et~al.]{TensorFlow}
Mart{\'\i}n Abadi, Paul Barham, Jianmin Chen, Zhifeng Chen, Andy Davis, Jeffrey
  Dean, Matthieu Devin, Sanjay Ghemawat, Geoffrey Irving, Michael Isard, et~al.
\newblock {TensorFlow}: a system for {Large-Scale} machine learning.
\newblock In \emph{12th USENIX symposium on operating systems design and
  implementation (OSDI 16)}, pp.\  265--283, 2016.

\bibitem[Ahmed et~al.(2021)Ahmed, Bakhtin, van~der Maaten, and
  Girdhar]{PHYREDynAware}
Eltayeb Ahmed, Anton Bakhtin, Laurens van~der Maaten, and Rohit Girdhar.
\newblock Physical reasoning using dynamics-aware models.
\newblock \emph{arXiv preprint arXiv:2102.10336}, 2021.

\bibitem[Ba et~al.(2016)Ba, Kiros, and Hinton]{LayerNorm}
Jimmy~Lei Ba, Jamie~Ryan Kiros, and Geoffrey~E Hinton.
\newblock Layer normalization.
\newblock \emph{arXiv preprint arXiv:1607.06450}, 2016.

\bibitem[Bakhtin et~al.(2019)Bakhtin, van~der Maaten, Johnson, Gustafson, and
  Girshick]{PHYRE}
Anton Bakhtin, Laurens van~der Maaten, Justin Johnson, Laura Gustafson, and
  Ross Girshick.
\newblock Phyre: A new benchmark for physical reasoning.
\newblock \emph{NeurIPS}, 32, 2019.

\bibitem[Battaglia et~al.(2016)Battaglia, Pascanu, Lai, Jimenez~Rezende,
  et~al.]{InteractionNetwork}
Peter Battaglia, Razvan Pascanu, Matthew Lai, Danilo Jimenez~Rezende, et~al.
\newblock Interaction networks for learning about objects, relations and
  physics.
\newblock \emph{NeurIPS}, 29, 2016.

\bibitem[Battaglia et~al.(2018)Battaglia, Hamrick, Bapst, Sanchez-Gonzalez,
  Zambaldi, Malinowski, Tacchetti, Raposo, Santoro, Faulkner,
  et~al.]{GNNSurvey}
Peter~W Battaglia, Jessica~B Hamrick, Victor Bapst, Alvaro Sanchez-Gonzalez,
  Vinicius Zambaldi, Mateusz Malinowski, Andrea Tacchetti, David Raposo, Adam
  Santoro, Ryan Faulkner, et~al.
\newblock Relational inductive biases, deep learning, and graph networks.
\newblock \emph{arXiv preprint arXiv:1806.01261}, 2018.

\bibitem[Bear et~al.(2021)Bear, Wang, Mrowca, Binder, Tung, Pramod, Holdaway,
  Tao, Smith, Sun, et~al.]{Physion}
Daniel Bear, Elias Wang, Damian Mrowca, Felix~Jedidja Binder, Hsiao-Yu Tung,
  RT~Pramod, Cameron Holdaway, Sirui Tao, Kevin~A Smith, Fan-Yun Sun, et~al.
\newblock Physion: Evaluating physical prediction from vision in humans and
  machines.
\newblock In \emph{NeurIPS Datasets and Benchmarks Track}, 2021.

\bibitem[Brown et~al.(2020)Brown, Mann, Ryder, Subbiah, Kaplan, Dhariwal,
  Neelakantan, Shyam, Sastry, Askell, et~al.]{GPT-3}
Tom Brown, Benjamin Mann, Nick Ryder, Melanie Subbiah, Jared~D Kaplan, Prafulla
  Dhariwal, Arvind Neelakantan, Pranav Shyam, Girish Sastry, Amanda Askell,
  et~al.
\newblock Language models are few-shot learners.
\newblock \emph{NeurIPS}, 33:\penalty0 1877--1901, 2020.

\bibitem[Burgess et~al.(2019)Burgess, Matthey, Watters, Kabra, Higgins,
  Botvinick, and Lerchner]{MONet}
Christopher~P Burgess, Loic Matthey, Nicholas Watters, Rishabh Kabra, Irina
  Higgins, Matt Botvinick, and Alexander Lerchner.
\newblock Monet: Unsupervised scene decomposition and representation.
\newblock \emph{arXiv preprint arXiv:1901.11390}, 2019.

\bibitem[Carion et~al.(2020)Carion, Massa, Synnaeve, Usunier, Kirillov, and
  Zagoruyko]{DETR}
Nicolas Carion, Francisco Massa, Gabriel Synnaeve, Nicolas Usunier, Alexander
  Kirillov, and Sergey Zagoruyko.
\newblock End-to-end object detection with transformers.
\newblock In \emph{ECCV}, pp.\  213--229. Springer, 2020.

\bibitem[Chang et~al.(2016)Chang, Ullman, Torralba, and
  Tenenbaum]{NeuralPhysicsEngine}
Michael Chang, Tomer Ullman, Antonio Torralba, and Joshua Tenenbaum.
\newblock A compositional object-based approach to learning physical dynamics.
\newblock In \emph{ICLR}, 2016.

\bibitem[Chen et~al.(2020{\natexlab{a}})Chen, Radford, Child, Wu, Jun, Luan,
  and Sutskever]{ImageGPT}
Mark Chen, Alec Radford, Rewon Child, Jeffrey Wu, Heewoo Jun, David Luan, and
  Ilya Sutskever.
\newblock Generative pretraining from pixels.
\newblock In \emph{ICML}, pp.\  1691--1703. PMLR, 2020{\natexlab{a}}.

\bibitem[Chen et~al.(2020{\natexlab{b}})Chen, Mao, Wu, Wong, Tenenbaum, and
  Gan]{DCL}
Zhenfang Chen, Jiayuan Mao, Jiajun Wu, Kwan-Yee~Kenneth Wong, Joshua~B
  Tenenbaum, and Chuang Gan.
\newblock Grounding physical concepts of objects and events through dynamic
  visual reasoning.
\newblock In \emph{ICLR}, 2020{\natexlab{b}}.

\bibitem[Creswell et~al.(2021)Creswell, Kabra, Burgess, and Shanahan]{OAT}
Antonia Creswell, Rishabh Kabra, Chris Burgess, and Murray Shanahan.
\newblock Unsupervised object-based transition models for 3d partially
  observable environments.
\newblock \emph{NeurIPS}, 34, 2021.

\bibitem[Deng et~al.(2009)Deng, Dong, Socher, Li, Li, and Fei-Fei]{ImageNet}
Jia Deng, Wei Dong, Richard Socher, Li-Jia Li, Kai Li, and Li~Fei-Fei.
\newblock Imagenet: A large-scale hierarchical image database.
\newblock In \emph{CVPR}, pp.\  248--255. IEEE, 2009.

\bibitem[Denton \& Fergus(2018)Denton and Fergus]{SVG-LP}
Emily Denton and Rob Fergus.
\newblock Stochastic video generation with a learned prior.
\newblock In \emph{ICML}, pp.\  1174--1183. PMLR, 2018.

\bibitem[Ding et~al.(2021{\natexlab{a}})Ding, Hill, Santoro, Reynolds, and
  Botvinick]{Aloe}
David Ding, Felix Hill, Adam Santoro, Malcolm Reynolds, and Matt Botvinick.
\newblock Attention over learned object embeddings enables complex visual
  reasoning.
\newblock \emph{NeurIPS}, 34, 2021{\natexlab{a}}.

\bibitem[Ding et~al.(2021{\natexlab{b}})Ding, Chen, Du, Luo, Tenenbaum, and
  Gan]{VRDP}
Mingyu Ding, Zhenfang Chen, Tao Du, Ping Luo, Josh Tenenbaum, and Chuang Gan.
\newblock Dynamic visual reasoning by learning differentiable physics models
  from video and language.
\newblock \emph{NeurIPS}, 34, 2021{\natexlab{b}}.

\bibitem[Dosovitskiy et~al.(2020)Dosovitskiy, Beyer, Kolesnikov, Weissenborn,
  Zhai, Unterthiner, Dehghani, Minderer, Heigold, Gelly, et~al.]{ViT}
Alexey Dosovitskiy, Lucas Beyer, Alexander Kolesnikov, Dirk Weissenborn,
  Xiaohua Zhai, Thomas Unterthiner, Mostafa Dehghani, Matthias Minderer, Georg
  Heigold, Sylvain Gelly, et~al.
\newblock An image is worth 16x16 words: Transformers for image recognition at
  scale.
\newblock In \emph{ICLR}, 2020.

\bibitem[Ebert et~al.(2017)Ebert, Finn, Lee, and Levine]{SNA}
Frederik Ebert, Chelsea Finn, Alex~X Lee, and Sergey Levine.
\newblock Self-supervised visual planning with temporal skip connections.
\newblock In \emph{CoRL}, pp.\  344--356, 2017.

\bibitem[Esser et~al.(2021)Esser, Rombach, and Ommer]{VQGAN}
Patrick Esser, Robin Rombach, and Bjorn Ommer.
\newblock Taming transformers for high-resolution image synthesis.
\newblock In \emph{CVPR}, pp.\  12873--12883, 2021.

\bibitem[Finn et~al.(2016)Finn, Goodfellow, and Levine]{CDNA}
Chelsea Finn, Ian Goodfellow, and Sergey Levine.
\newblock Unsupervised learning for physical interaction through video
  prediction.
\newblock \emph{NeurIPS}, 29, 2016.

\bibitem[Fragkiadaki et~al.(2016)Fragkiadaki, Agrawal, Levine, and
  Malik]{BilliardsVP}
Katerina Fragkiadaki, Pulkit Agrawal, Sergey Levine, and Jitendra Malik.
\newblock Learning visual predictive models of physics for playing billiards.
\newblock In \emph{ICLR}, 2016.

\bibitem[Girdhar et~al.(2020)Girdhar, Gustafson, Adcock, and van~der
  Maaten]{PHYREDecBaseline}
Rohit Girdhar, Laura Gustafson, Aaron Adcock, and Laurens van~der Maaten.
\newblock Forward prediction for physical reasoning.
\newblock \emph{arXiv preprint arXiv:2006.10734}, 2020.

\bibitem[Girshick(2015)]{FastRCNN}
Ross Girshick.
\newblock Fast r-cnn.
\newblock In \emph{ICCV}, pp.\  1440--1448, 2015.

\bibitem[Hochreiter \& Schmidhuber(1997)Hochreiter and Schmidhuber]{LSTM}
Sepp Hochreiter and J{\"u}rgen Schmidhuber.
\newblock Long short-term memory.
\newblock \emph{Neural computation}, 9\penalty0 (8):\penalty0 1735--1780, 1997.

\bibitem[Jiang et~al.(2019)Jiang, Janghorbani, De~Melo, and Ahn]{SCALOR}
Jindong Jiang, Sepehr Janghorbani, Gerard De~Melo, and Sungjin Ahn.
\newblock Scalor: Generative world models with scalable object representations.
\newblock In \emph{ICLR}, 2019.

\bibitem[Johnson et~al.(2017)Johnson, Hariharan, Van Der~Maaten, Fei-Fei,
  Lawrence~Zitnick, and Girshick]{CLEVR}
Justin Johnson, Bharath Hariharan, Laurens Van Der~Maaten, Li~Fei-Fei,
  C~Lawrence~Zitnick, and Ross Girshick.
\newblock Clevr: A diagnostic dataset for compositional language and elementary
  visual reasoning.
\newblock In \emph{CVPR}, pp.\  2901--2910, 2017.

\bibitem[Kabra et~al.(2021)Kabra, Zoran, Erdogan, Matthey, Creswell, Botvinick,
  Lerchner, and Burgess]{SIMONe}
Rishabh Kabra, Daniel Zoran, Goker Erdogan, Loic Matthey, Antonia Creswell,
  Matt Botvinick, Alexander Lerchner, and Chris Burgess.
\newblock Simone: View-invariant, temporally-abstracted object representations
  via unsupervised video decomposition.
\newblock \emph{NeurIPS}, 34, 2021.

\bibitem[Kenton \& Toutanova(2019)Kenton and Toutanova]{BERT}
Jacob Devlin Ming-Wei~Chang Kenton and Lee~Kristina Toutanova.
\newblock Bert: Pre-training of deep bidirectional transformers for language
  understanding.
\newblock In \emph{NAACL}, pp.\  4171--4186, 2019.

\bibitem[Kingma \& Ba(2015)Kingma and Ba]{Adam}
Diederik~P Kingma and Jimmy Ba.
\newblock Adam: A method for stochastic optimization.
\newblock In \emph{ICLR}, 2015.

\bibitem[Kipf et~al.(2019)Kipf, van~der Pol, and Welling]{CSWM}
Thomas Kipf, Elise van~der Pol, and Max Welling.
\newblock Contrastive learning of structured world models.
\newblock \emph{arXiv preprint arXiv:1911.12247}, 2019.

\bibitem[Kipf et~al.(2021)Kipf, Elsayed, Mahendran, Stone, Sabour, Heigold,
  Jonschkowski, Dosovitskiy, and Greff]{SAVi}
Thomas Kipf, Gamaleldin~F Elsayed, Aravindh Mahendran, Austin Stone, Sara
  Sabour, Georg Heigold, Rico Jonschkowski, Alexey Dosovitskiy, and Klaus
  Greff.
\newblock Conditional object-centric learning from video.
\newblock \emph{arXiv preprint arXiv:2111.12594}, 2021.

\bibitem[Kosiorek et~al.(2018)Kosiorek, Kim, Teh, and Posner]{SQAIR}
Adam Kosiorek, Hyunjik Kim, Yee~Whye Teh, and Ingmar Posner.
\newblock Sequential attend, infer, repeat: Generative modelling of moving
  objects.
\newblock \emph{NeurIPS}, 31, 2018.

\bibitem[Kossen et~al.(2019)Kossen, Stelzner, Hussing, Voelcker, and
  Kersting]{STOVE}
Jannik Kossen, Karl Stelzner, Marcel Hussing, Claas Voelcker, and Kristian
  Kersting.
\newblock Structured object-aware physics prediction for video modeling and
  planning.
\newblock In \emph{ICLR}, 2019.

\bibitem[Lee et~al.(2018)Lee, Zhang, Ebert, Abbeel, Finn, and Levine]{SAVP}
Alex~X Lee, Richard Zhang, Frederik Ebert, Pieter Abbeel, Chelsea Finn, and
  Sergey Levine.
\newblock Stochastic adversarial video prediction.
\newblock \emph{arXiv preprint arXiv:1804.01523}, 2018.

\bibitem[Li et~al.(2019)Li, Wu, Zhu, Tenenbaum, Torralba, and Tedrake]{PropNet}
Yunzhu Li, Jiajun Wu, Jun-Yan Zhu, Joshua~B Tenenbaum, Antonio Torralba, and
  Russ Tedrake.
\newblock Propagation networks for model-based control under partial
  observation.
\newblock In \emph{ICRA}, pp.\  1205--1211. IEEE, 2019.

\bibitem[Lin et~al.(2019)Lin, Wu, Peri, Sun, Singh, Deng, Jiang, and
  Ahn]{SPACE}
Zhixuan Lin, Yi-Fu Wu, Skand~Vishwanath Peri, Weihao Sun, Gautam Singh, Fei
  Deng, Jindong Jiang, and Sungjin Ahn.
\newblock Space: Unsupervised object-oriented scene representation via spatial
  attention and decomposition.
\newblock In \emph{ICLR}, 2019.

\bibitem[Lin et~al.(2020)Lin, Wu, Peri, Fu, Jiang, and Ahn]{G-SWM}
Zhixuan Lin, Yi-Fu Wu, Skand Peri, Bofeng Fu, Jindong Jiang, and Sungjin Ahn.
\newblock Improving generative imagination in object-centric world models.
\newblock In \emph{ICML}, pp.\  6140--6149. PMLR, 2020.

\bibitem[Liu et~al.(2021)Liu, Lin, Cao, Hu, Wei, Zhang, Lin, and
  Guo]{SwinTransformer}
Ze~Liu, Yutong Lin, Yue Cao, Han Hu, Yixuan Wei, Zheng Zhang, Stephen Lin, and
  Baining Guo.
\newblock Swin transformer: Hierarchical vision transformer using shifted
  windows.
\newblock In \emph{ICCV}, pp.\  10012--10022, 2021.

\bibitem[Locatello et~al.(2020)Locatello, Weissenborn, Unterthiner, Mahendran,
  Heigold, Uszkoreit, Dosovitskiy, and Kipf]{SlotAttn}
Francesco Locatello, Dirk Weissenborn, Thomas Unterthiner, Aravindh Mahendran,
  Georg Heigold, Jakob Uszkoreit, Alexey Dosovitskiy, and Thomas Kipf.
\newblock Object-centric learning with slot attention.
\newblock \emph{NeurIPS}, 33:\penalty0 11525--11538, 2020.

\bibitem[Micheli et~al.(2022)Micheli, Alonso, and
  Fleuret]{TransformerWorldModel}
Vincent Micheli, Eloi Alonso, and Francois Fleuret.
\newblock Transformers are sample efficient world models.
\newblock \emph{arXiv preprint arXiv:2209.00588}, 2022.

\bibitem[Nash et~al.(2022)Nash, Carreira, Walker, Barr, Jaegle, Malinowski, and
  Battaglia]{Transframer}
Charlie Nash, Jo{\~a}o Carreira, Jacob Walker, Iain Barr, Andrew Jaegle,
  Mateusz Malinowski, and Peter Battaglia.
\newblock Transframer: Arbitrary frame prediction with generative models.
\newblock \emph{arXiv preprint arXiv:2203.09494}, 2022.

\bibitem[Oprea et~al.(2020)Oprea, Martinez-Gonzalez, Garcia-Garcia,
  Castro-Vargas, Orts-Escolano, Garcia-Rodriguez, and Argyros]{VPReview}
Sergiu Oprea, Pablo Martinez-Gonzalez, Alberto Garcia-Garcia, John~Alejandro
  Castro-Vargas, Sergio Orts-Escolano, Jose Garcia-Rodriguez, and Antonis
  Argyros.
\newblock A review on deep learning techniques for video prediction.
\newblock \emph{TPAMI}, 2020.

\bibitem[Paszke et~al.(2019)Paszke, Gross, Massa, Lerer, Bradbury, Chanan,
  Killeen, Lin, Gimelshein, Antiga, et~al.]{PyTorch}
Adam Paszke, Sam Gross, Francisco Massa, Adam Lerer, James Bradbury, Gregory
  Chanan, Trevor Killeen, Zeming Lin, Natalia Gimelshein, Luca Antiga, et~al.
\newblock Pytorch: An imperative style, high-performance deep learning library.
\newblock \emph{NeurIPS}, 32, 2019.

\bibitem[Qi et~al.(2017)Qi, Su, Mo, and Guibas]{PointNet}
Charles~R Qi, Hao Su, Kaichun Mo, and Leonidas~J Guibas.
\newblock Pointnet: Deep learning on point sets for 3d classification and
  segmentation.
\newblock In \emph{CVPR}, pp.\  652--660, 2017.

\bibitem[Qi et~al.(2020)Qi, Wang, Pathak, Ma, and Malik]{RPIN}
Haozhi Qi, Xiaolong Wang, Deepak Pathak, Yi~Ma, and Jitendra Malik.
\newblock Learning long-term visual dynamics with region proposal interaction
  networks.
\newblock In \emph{ICLR}, 2020.

\bibitem[Radford et~al.(2018)Radford, Narasimhan, Salimans, and
  Sutskever]{GPT-1}
Alec Radford, Karthik Narasimhan, Tim Salimans, and Ilya Sutskever.
\newblock Improving language understanding by generative pre-training.
\newblock 2018.

\bibitem[Radford et~al.(2019)Radford, Wu, Child, Luan, Amodei, Sutskever,
  et~al.]{GPT-2}
Alec Radford, Jeffrey Wu, Rewon Child, David Luan, Dario Amodei, Ilya
  Sutskever, et~al.
\newblock Language models are unsupervised multitask learners.
\newblock \emph{OpenAI blog}, 1\penalty0 (8):\penalty0 9, 2019.

\bibitem[Razavi et~al.(2019)Razavi, Van~den Oord, and Vinyals]{VQVAE}
Ali Razavi, Aaron Van~den Oord, and Oriol Vinyals.
\newblock Generating diverse high-fidelity images with vq-vae-2.
\newblock \emph{NeurIPS}, 32, 2019.

\bibitem[Ren \& Wang(2022)Ren and Wang]{lookoutside}
Xuanchi Ren and Xiaolong Wang.
\newblock Look outside the room: Synthesizing a consistent long-term 3d scene
  video from a single image.
\newblock \emph{arXiv preprint arXiv:2203.09457}, 2022.

\bibitem[Sanchez-Gonzalez et~al.(2018)Sanchez-Gonzalez, Heess, Springenberg,
  Merel, Riedmiller, Hadsell, and Battaglia]{GNNPhysicsEngine}
Alvaro Sanchez-Gonzalez, Nicolas Heess, Jost~Tobias Springenberg, Josh Merel,
  Martin Riedmiller, Raia Hadsell, and Peter Battaglia.
\newblock Graph networks as learnable physics engines for inference and
  control.
\newblock In \emph{ICML}, pp.\  4470--4479. PMLR, 2018.

\bibitem[Santoro et~al.(2017)Santoro, Raposo, Barrett, Malinowski, Pascanu,
  Battaglia, and Lillicrap]{RelationNetwork}
Adam Santoro, David Raposo, David~G Barrett, Mateusz Malinowski, Razvan
  Pascanu, Peter Battaglia, and Timothy Lillicrap.
\newblock A simple neural network module for relational reasoning.
\newblock \emph{NeurIPS}, 30, 2017.

\bibitem[Sara et~al.(2019)Sara, Akter, and Uddin]{ImageQuality}
Umme Sara, Morium Akter, and Mohammad~Shorif Uddin.
\newblock Image quality assessment through fsim, ssim, mse and psnr—a
  comparative study.
\newblock \emph{Journal of Computer and Communications}, 7\penalty0
  (3):\penalty0 8--18, 2019.

\bibitem[Shi et~al.(2015)Shi, Chen, Wang, Yeung, Wong, and Woo]{ConvLSTM}
Xingjian Shi, Zhourong Chen, Hao Wang, Dit-Yan Yeung, Wai-Kin Wong, and
  Wang-chun Woo.
\newblock Convolutional lstm network: A machine learning approach for
  precipitation nowcasting.
\newblock \emph{NeurIPS}, 28, 2015.

\bibitem[Singh et~al.(2021)Singh, Deng, and Ahn]{SLATE}
Gautam Singh, Fei Deng, and Sungjin Ahn.
\newblock Illiterate dall-e learns to compose.
\newblock In \emph{ICLR}, 2021.

\bibitem[Singh et~al.(2022)Singh, Wu, and Ahn]{STEVE}
Gautam Singh, Yi-Fu Wu, and Sungjin Ahn.
\newblock Simple unsupervised object-centric learning for complex and
  naturalistic videos.
\newblock \emph{arXiv preprint arXiv:2205.14065}, 2022.

\bibitem[Stelzner et~al.(2019)Stelzner, Peharz, and Kersting]{SuPAIR}
Karl Stelzner, Robert Peharz, and Kristian Kersting.
\newblock Faster attend-infer-repeat with tractable probabilistic models.
\newblock In \emph{ICML}, pp.\  5966--5975. PMLR, 2019.

\bibitem[Sun et~al.(2022)Sun, Huang, Lu, Liu, Zhou, and Garg]{sun2022plate}
Jiankai Sun, De-An Huang, Bo~Lu, Yun-Hui Liu, Bolei Zhou, and Animesh Garg.
\newblock Plate: Visually-grounded planning with transformers in procedural
  tasks.
\newblock \emph{IEEE Robotics and Automation Letters}, 7\penalty0 (2):\penalty0
  4924--4930, 2022.

\bibitem[Tevet et~al.(2022)Tevet, Gordon, Hertz, Bermano, and
  Cohen-Or]{MotionClip}
Guy Tevet, Brian Gordon, Amir Hertz, Amit~H Bermano, and Daniel Cohen-Or.
\newblock Motionclip: Exposing human motion generation to clip space.
\newblock \emph{arXiv preprint arXiv:2203.08063}, 2022.

\bibitem[Touvron et~al.(2021)Touvron, Cord, Douze, Massa, Sablayrolles, and
  J{\'e}gou]{DEIT}
Hugo Touvron, Matthieu Cord, Matthijs Douze, Francisco Massa, Alexandre
  Sablayrolles, and Herv{\'e} J{\'e}gou.
\newblock Training data-efficient image transformers \& distillation through
  attention.
\newblock In \emph{ICML}, pp.\  10347--10357. PMLR, 2021.

\bibitem[van Steenkiste et~al.(2018)van Steenkiste, Chang, Greff, and
  Schmidhuber]{R-NEM}
Sjoerd van Steenkiste, Michael Chang, Klaus Greff, and J{\"u}rgen Schmidhuber.
\newblock Relational neural expectation maximization: Unsupervised discovery of
  objects and their interactions.
\newblock In \emph{ICLR}, 2018.

\bibitem[Vaswani et~al.(2017)Vaswani, Shazeer, Parmar, Uszkoreit, Jones, Gomez,
  Kaiser, and Polosukhin]{Attention}
Ashish Vaswani, Noam Shazeer, Niki Parmar, Jakob Uszkoreit, Llion Jones,
  Aidan~N Gomez, Lukasz Kaiser, and Illia Polosukhin.
\newblock Attention is all you need.
\newblock \emph{NeurIPS}, 30, 2017.

\bibitem[Veerapaneni et~al.(2020)Veerapaneni, Co-Reyes, Chang, Janner, Finn,
  Wu, Tenenbaum, and Levine]{OP3}
Rishi Veerapaneni, John~D Co-Reyes, Michael Chang, Michael Janner, Chelsea
  Finn, Jiajun Wu, Joshua Tenenbaum, and Sergey Levine.
\newblock Entity abstraction in visual model-based reinforcement learning.
\newblock In \emph{CoRL}, pp.\  1439--1456. PMLR, 2020.

\bibitem[Villegas et~al.(2017)Villegas, Yang, Zou, Sohn, Lin, and Lee]{posevp1}
Ruben Villegas, Jimei Yang, Yuliang Zou, Sungryull Sohn, Xunyu Lin, and Honglak
  Lee.
\newblock Learning to generate long-term future via hierarchical prediction.
\newblock In \emph{ICML}, pp.\  3560--3569. PMLR, 2017.

\bibitem[Walker et~al.(2016)Walker, Doersch, Gupta, and Hebert]{flowvp1}
Jacob Walker, Carl Doersch, Abhinav Gupta, and Martial Hebert.
\newblock An uncertain future: Forecasting from static images using variational
  autoencoders.
\newblock In \emph{ECCV}, pp.\  835--851. Springer, 2016.

\bibitem[Wang et~al.(2017)Wang, Long, Wang, Gao, and Yu]{PredRNN}
Yunbo Wang, Mingsheng Long, Jianmin Wang, Zhifeng Gao, and Philip~S Yu.
\newblock Predrnn: Recurrent neural networks for predictive learning using
  spatiotemporal lstms.
\newblock \emph{NeurIPS}, 30, 2017.

\bibitem[Wang et~al.(2004)Wang, Bovik, Sheikh, and Simoncelli]{SSIM}
Zhou Wang, Alan~C Bovik, Hamid~R Sheikh, and Eero~P Simoncelli.
\newblock Image quality assessment: from error visibility to structural
  similarity.
\newblock \emph{TIP}, 13\penalty0 (4):\penalty0 600--612, 2004.

\bibitem[Watters et~al.(2017)Watters, Zoran, Weber, Battaglia, Pascanu, and
  Tacchetti]{VIN}
Nicholas Watters, Daniel Zoran, Theophane Weber, Peter Battaglia, Razvan
  Pascanu, and Andrea Tacchetti.
\newblock Visual interaction networks: Learning a physics simulator from video.
\newblock \emph{NeurIPS}, 30, 2017.

\bibitem[Wightman(2019)]{TIMM}
Ross Wightman.
\newblock Pytorch image models.
\newblock \url{https://github.com/rwightman/pytorch-image-models}, 2019.

\bibitem[Wu et~al.(2015)Wu, Yildirim, Lim, Freeman, and Tenenbaum]{Galileo}
Jiajun Wu, Ilker Yildirim, Joseph~J Lim, Bill Freeman, and Josh Tenenbaum.
\newblock Galileo: Perceiving physical object properties by integrating a
  physics engine with deep learning.
\newblock \emph{NeurIPS}, 28, 2015.

\bibitem[Wu et~al.(2016)Wu, Lim, Zhang, Tenenbaum, and Freeman]{Physics101}
Jiajun Wu, Joseph~J Lim, Hongyi Zhang, Joshua~B Tenenbaum, and William~T
  Freeman.
\newblock Physics 101: Learning physical object properties from unlabeled
  videos.
\newblock In \emph{BMVC}, volume~2, pp.\ ~7, 2016.

\bibitem[Wu et~al.(2021)Wu, Yoon, and Ahn]{OCVT}
Yi-Fu Wu, Jaesik Yoon, and Sungjin Ahn.
\newblock Generative video transformer: Can objects be the words?
\newblock In \emph{ICML}, pp.\  11307--11318. PMLR, 2021.

\bibitem[Xiong et~al.(2020)Xiong, Yang, He, Zheng, Zheng, Xing, Zhang, Lan,
  Wang, and Liu]{Pre-LN-Transformer}
Ruibin Xiong, Yunchang Yang, Di~He, Kai Zheng, Shuxin Zheng, Chen Xing,
  Huishuai Zhang, Yanyan Lan, Liwei Wang, and Tieyan Liu.
\newblock On layer normalization in the transformer architecture.
\newblock In \emph{ICML}, pp.\  10524--10533. PMLR, 2020.

\bibitem[Yan et~al.(2021)Yan, Zhang, Abbeel, and Srinivas]{VideoGPT}
Wilson Yan, Yunzhi Zhang, Pieter Abbeel, and Aravind Srinivas.
\newblock Videogpt: Video generation using vq-vae and transformers.
\newblock \emph{arXiv preprint arXiv:2104.10157}, 2021.

\bibitem[Yang et~al.(2021)Yang, Lamdouar, Lu, Zisserman, and
  Xie]{MotionGroupingVOS}
Charig Yang, Hala Lamdouar, Erika Lu, Andrew Zisserman, and Weidi Xie.
\newblock Self-supervised video object segmentation by motion grouping.
\newblock In \emph{ICCV}, pp.\  7177--7188, 2021.

\bibitem[Ye et~al.(2019)Ye, Singh, Gupta, and Tulsiani]{CVP}
Yufei Ye, Maneesh Singh, Abhinav Gupta, and Shubham Tulsiani.
\newblock Compositional video prediction.
\newblock In \emph{ICCV}, pp.\  10353--10362, 2019.

\bibitem[Yi et~al.(2019)Yi, Gan, Li, Kohli, Wu, Torralba, and
  Tenenbaum]{CLEVRER}
Kexin Yi, Chuang Gan, Yunzhu Li, Pushmeet Kohli, Jiajun Wu, Antonio Torralba,
  and Joshua~B Tenenbaum.
\newblock Clevrer: Collision events for video representation and reasoning.
\newblock In \emph{ICLR}, 2019.

\bibitem[Yu et~al.(2022)Yu, Chen, Yin, Easterbrook, and Garg]{yu2022mac}
Wei Yu, Wenxin Chen, Songheng Yin, Steve Easterbrook, and Animesh Garg.
\newblock Modular action concept grounding in semantic video prediction.
\newblock In \emph{CVPR}, pp.\  3605--3614, 2022.

\bibitem[Zhang et~al.(2018)Zhang, Isola, Efros, Shechtman, and Wang]{LPIPS}
Richard Zhang, Phillip Isola, Alexei~A Efros, Eli Shechtman, and Oliver Wang.
\newblock The unreasonable effectiveness of deep features as a perceptual
  metric.
\newblock In \emph{CVPR}, pp.\  586--595, 2018.

\bibitem[Zoran et~al.(2021)Zoran, Kabra, Lerchner, and Rezende]{PARTS}
Daniel Zoran, Rishabh Kabra, Alexander Lerchner, and Danilo~J Rezende.
\newblock Parts: Unsupervised segmentation with slots, attention and
  independence maximization.
\newblock In \emph{ICCV}, pp.\  10439--10447, 2021.

\end{thebibliography}
\bibliographystyle{iclr2023_conference}

\clearpage

\appendix

\section{Additional Related Work}\label{appendix:related-work}


\heading{Physical reasoning for dynamics modeling.}
Instead of explicitly encoding physical laws to deep models and estimating latent variables from inputs~\citep{Galileo,Physics101}, recent approaches implicitly infer them by modeling object interactions in the scene~\citep{NeuralPhysicsEngine,GNNSurvey,InteractionNetwork,GNNPhysicsEngine,PropNet}.
VIN~\citep{VIN} employs a CNN to encode video frames into a multi-channel 1D tensor, where each channel represents an object.
It enforces a fixed mapping between object identity and feature channel, which cannot generalize to different number of objects with varying appearance.
CVP~\citep{CVP} leverage object bounding boxes to crop input image and apply CNNs to extract object-centric representations.
Since object features are extracted from raw image patches, it ignores the context information and thus cannot model the interactions between objects and the environment.
RPIN~\citep{RPIN} instead uses RoIPooling~\citep{FastRCNN} to extract object features from image feature maps, which contains background information.
However, these methods rely on ground-truth object-level annotations for training, which are often unavailable.
In contrast, \algoNameFull pre-trains unsupervised object-centric models on unlabeled videos.
It ensures accurate decomposition of foreground objects and background environment, laying the foundation for building powerful dynamics models.

\heading{Dynamics modeling in object-centric representation learning.}
R-NEM~\citep{R-NEM} is the first end-to-end object-centric model to reason about objects and their interactions from pixel observations alone.
It extracts object features from raw observations and uses an interaction function in the form of a GNN to model interactions.
SCALOR~\citep{SCALOR} scales the SQAIR~\citep{SQAIR} model to work on scenes with multiple moving objects.
It introduces a background module to model the image background separately.
It also equips each object with a depth property to handle occlusions.
STOVE~\citep{STOVE} incorporates a GNN-based dynamics model into SuPAIR~\citep{SuPAIR} to reason object interactions, where object representations are explicitly disentangled into positions, velocities and appearance.
Similarly, OP3~\citep{OP3} learns pairwise relationship between objects based on a symmetric assumption.
G-SWM~\citep{G-SWM} combines the key properties of the above methods and proposes a unified framework for accurate dynamics prediction.
A hierarchical latent modeling technique is utilized to handle the multi-modality of the scene dynamics.
Leveraging the power of Transformers, OAT~\citep{OAT} directly learns to align slots extracted from each frame to gain temporal consistency and perform slot interactions.
However, the temporal dynamics is still modeled by an LSTM~\citep{LSTM} module.
Similarly, PARTS~\citep{PARTS} employs the same Transformer-LSTM module from OAT.
It utilizes the Slot-Attention~\citep{SlotAttn} mechanism to detect objects and relies on a fixed independent prior to achieve stable future rollout performance.
OCVT~\citep{OCVT} is the most relevant work to \algoNameFull.
It also applies Transformer over slots from multiple frames and performs future prediction in an autoregressive manner.
However, OCVT still disentangles its underlying object features into position, depth and semantic information.
It also relies on a Hungarian matching algorithm to achieve temporal alignment of slots.
As a result, OCVT is inferior to G-SWM in terms of future rollout.
Compared to previous works, \algoNameFull is a general Transformer-based dynamics model that is agnostic to the object-centric representations it builds upon.
It does not assume any explicit disentanglement of the object property, while still can handle the object interactions well.
Without the use of RNNs or GNNs, we achieve state-of-the-art dynamics modeling ability.

\heading{Transformers.}
With the prevalence of Transformers in the NLP field~\citep{Attention,BERT}, there have been tremendous efforts in introducing it to computer vision tasks~\citep{ViT,DETR,SwinTransformer}.
Our method is highly motivated by previous works in Transformer-based autoregressive image and video generation~\citep{VQGAN,ImageGPT,VideoGPT,Transframer,lookoutside}.
VQGAN~\citep{VQGAN} first pretrains the encoder, decoder and a codebook that can map images to discrete tokens and tokens back to images.
Then, a GPT-like Transformer model is trained to autoregressively predict the input tokens for high-fidelity image generation.
Transframer~\citep{Transframer} instead discretizes video frames using Discrete Cosine Transform (DCT), and learns an autoregressive Transformer over these sparse representations from multiple frames.
The design of \algoNameFull is mostly related to~\citep{lookoutside}, which also uses image tokens from multiple frames to enable consistent long-term view synthesis.
Different from these works, our mapping step maps images to object-centric representations, preserving the identity of objects and is independent of the input image resolution.

\section{Dataset Details}\label{appendix:dataset}

\heading{OBJ3D~\citep{G-SWM}.}
The videos in this dataset are generated by first placing 3 to 5 static objects in the scene, and then launching a sphere from the front of the scene to collide with those objects.
Compared to CLEVRER, the objects in OBJ3D occupy more pixels in images, have less collisions and occlusions, and are all visible in the scene at the beginning of the videos.

\heading{CLEVRER~\citep{CLEVRER}.}
The videos in this dataset contain static or moving objects at the beginning, and there will be various new objects entering the scene from random directions throughout the video.
The smaller size and more diverse interactions of objects make CLEVRER more challenging than OBJ3D.
We obtain the ground-truth segmentation masks from their official website, which are used to generate object bounding boxes.
We calculate the Average Recall (AR) with an IoU threshold of $50\%$ for the predicted object boxes and the Adjusted Rand Index (ARI) for the object masks.
We also report a variant of ARI and mIoU which only focus on foreground objects termed FG-ARI and FG-mIoU as done in the SAVi paper~\citep{SAVi}.

As a Visual Question Answering (VQA) dataset, CLEVRER consists of four types of questions generated by template-based programs, namely, descriptive, explanatory, predictive and counterfactual.
The latter three types of questions are multiple-choice questions, where the VQA model needs to classify whether each choice is correct.

\heading{Physion~\citep{Physion}.}
This dataset contains eight physical scenarios, each falls under a common physical phenomenon, such as rigid- and soft-body collisions, falling, rolling and sliding motions.
The foreground objects used in the simulation vary in categories, textures, colors and sizes.
It also uses diverse background as the scene environment, and randomize the camera pose in rendering the videos.
Overall, this dataset presents more complex visual appearance and object dynamics compared to other synthetic VQA datasets.
Therefore, we apply the recently proposed powerful object-centric model, STEVE~\citep{STEVE}, to extract slots on this dataset.

Physion splits the videos into three sets, namely, \textit{Training}, \textit{Readout Fitting} and \textit{Testing}.
We truncate all videos by 150 frames as most of the interactions end before that, and sub-sample the videos by a factor of 3 for training the dynamics model.
Following the official evaluation protocol, the dynamics models are first trained on videos from the Training set under future prediction loss.
Then, they observe the first 45 frames of videos in the Readout Fitting and Testing set, and perform rollout to generate future scene representations (e.g. feature maps for image-based dynamics models, or object slots for \algoNameFull).
A linear readout model is trained on observed and rollout scene representations from the Readout Fitting set to classify whether the two cued objects (one in red and one in yellow) contact.
Finally, the classification accuracy of the trained readout model on the Testing set scene representations is reported.
Please refer to their paper~\citep{Physion} for detailed description of the evaluation.

\heading{PHYRE~\citep{PHYRE}.}\label{appendix-phyre-dataset}
We study the PHYRE-B tier in this paper, which consists of 25 templates of tasks.
Tasks within the same template share similar initial configuration of the objects.
There are two evaluation settings, namely, \textit{within-template}, where training and testing tasks come from the same templates, and \textit{cross-template}, where train-test tasks are from different templates.
We simulate the videos in 1 FPS as done in previous works~\citep{PHYRE,RPIN}.

In our preliminary experiment, we discovered that SAVi usually fails to detect objects in light color (e.g. the green and gray balls).
We hypothesize that this is because the $L_2$ norm of light colors is close to white (the color of background), so pixels in light colors receive gradients of small magnitude during optimization, leading to worse segmentation results.
Since improving object-centric models is not the focus of this paper, we choose a simple workaround by changing the background color from white to black.

To solve a task at test time, models need to determine the size of a red ball and the location to put it in the scene, such that the green ball touches the blue/purple object for more than 3 seconds after the scene evolves.
Following~\citep{RPIN}, we train models to score a pre-defined 10,000 actions when applied to the current task (by rendering the red ball in the scene), and execute them according to the ranking.
The evaluation metric, \textit{AUCCESS}, is a weighted sum of the Area Under Curve (AUC) of the success percentage vs number of attempts curve for the first 100 attempts.
See their paper~\citep{PHYRE} for detailed explanation of the metrics.

\section{Implementation Details}\label{appendix:implement-ours}

We provide more implementation details of our method in this section.
Table~\ref{appendix-table:implementation} lists the hyper-parameters used in our experiments to facilitate reproduction.

\heading{SAVi.}
We reproduce the unconditional version of SAVi in PyTorch~\citep{PyTorch} to perform unsupervised object discovery.
Specifically, we use the same CNN encoder, decoder, Slot Attention based corrector and Transformer based predictor as their experiments on CATER, except on PHYRE the spatial broadcast size of the decoder is $16 \times 16$ to better capture small objects.
The slot size is $128$ and the training video clip length is 6 on all the datasets.
We adopt Adam~\citep{Adam} as the training optimizer.
We use the same warmup and decay learning rate schedule which first linearly increases from $0$ to $2 \times 10^{-4}$ for the first $2.5\%$ of the total training steps, and then decrease to $0$ in a cosine annealing strategy.
We perform gradient clipping with a maximum norm of $0.05$.

\begin{table}[t]
    \vspace{\pagetopmargin}
    \centering
    \small
    \begin{tabular}{cccccc}
        \toprule
        \multicolumn{2}{c}{\textbf{Dataset}} & OBJ3D & CLEVRER & Physion & PHYRE \\
        \midrule
        \multirow{6}{*}{Slot Model} & Base Model & SAVi & SAVi & STEVE & SAVi \\
        & Image Resolution & $64 \times 64$ & $64 \times 64$ & $128 \times 128$ & $128 \times 128$ \\
        & Number of Slots $N$ & 6 & 7 & 6 & 8 \\
        & Slot Size $D_{slot}$ & 128 & 128 & 192 & 128 \\
        & Batch Size & 64 & 64 & 48 & 64 \\
        & Training Steps & 80k & 200k & 460k & 370k \\
        \midrule
        \multirow{7}{*}{Transformer} & Burn-in Steps $T$ & 6 & 6 & 15 & 1 \\
        & Rollout Steps $K$ & 10 & 10 & 10 & 10 \\
        & Latent Size $D_e$ & 128 & 256 & 256 & 256 \\
        & Number of Layers $N_\mathcal{T}$ & 4 & 4 & 8 & 8 \\
        & Loss Weight $\lambda$ & 1 & 1 & 0 & 0 \\
        & Batch Size & 128 & 128 & 128 & 64 \\
        & Training Steps & 200k & 500k & 250k & 300k \\
        \bottomrule
    \end{tabular}
    \caption{
    Variations in model architectures and training settings on different datasets.
    \label{appendix-table:implementation}
    }
    \vspace{\tablemargin}
\end{table}

\begin{figure}[t]
    \centering
    \vspace{-2mm}
    \hspace{-10mm}
    \subfigure{
        \includegraphics[width=0.85\textwidth]{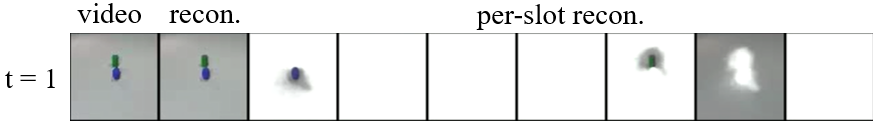}
    }\\
    \vspace{-2.5mm}
    \hspace{-10mm}
    \subfigure{
        \includegraphics[width=0.85\textwidth]{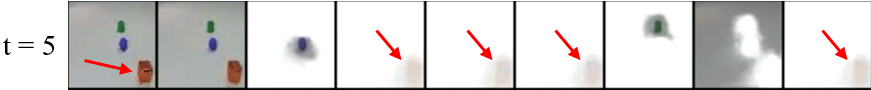}
    }\\
    \vspace{-2.5mm}
    \hspace{-10.2mm}
    \subfigure{
        \includegraphics[width=0.85\textwidth]{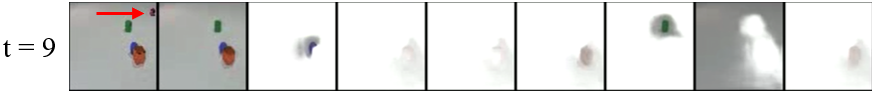}
    }
    \caption{
    Illustration for missing objects of vanilla SAVi on CLEVRER videos.
    There are two objects at the beginning of this video (top).
    When the red cube enters the scene, all 4 empty slots attend to this object, resulting in object sharing (middle).
    When another object enters the scene from the top right corner, SAVi does not have empty slots to detect it (bottom).
    As a result, this object is ignored by the model.
    }
    \label{appendix-fig:vanilla-savi-clevrer}
    \vspace{-1mm}
\end{figure}

\heading{Stochastic SAVi.}
As stated in the main paper, vanilla SAVi sometimes fails to capture newly entered objects in a video, and we detail the reason and our solution as follows.
We use 7 slots for SAVi on CLEVRER which has a maximum of 6 objects in the scene.
Imagine a video with 4 objects $\{O_i\}_{i=1}^4$ at the beginning.
Let us assume SAVi captures the objects in the first 4 slots and the background in the 5th slot.
This leads to two empty slots $\bm{s}_6$ and $\bm{s}_7$, which are very similar with L2 distance $||\bm{s}_6 - \bm{s}_7||^2$ generally smaller than $0.05$.
Consequently, when there is a new object $O_5$ enters the scene, $\bm{s}_6$ and $\bm{s}_7$ will both attend to it, resulting in object sharing between slots.
Now, if there is another object $O_6$ entering the scene, there will be no "free" slot to detect this new object.
Therefore, $O_6$ will be ignored by SAVi, until one of the previous object leaves the scene.
An example is shown in Figure~\ref{appendix-fig:vanilla-savi-clevrer}.
This issue occurs only on CLEVRER because all the objects are presented in videos from the beginning in other datasets.
Besides, SAVi did not experiment on datasets with multiple newly entered objects \footnote{confirmed with the authors of SAVi}, and thus they did not observe such problem.

From our analysis, the issue stems from the similarity of empty slots, which is because of the permutation equivariance of slots.
To break the symmetry, we introduce stochasticity to slots initialized from previous timestep.
Specifically, we modify the slot transition function by applying a two-layer MLP with Layer Normalization~\citep{LayerNorm} to predict the mean and log variance of $\tilde{\mathcal{S}}_{t+1}$:
\begin{equation}
    (\mu_{t+1}, \log \sigma^2_{t+1}) = \mathrm{MLP}(f_{trans}(\mathcal{S}_t)).
\end{equation}
Then, we sample from this distribution to get $\tilde{\mathcal{S}}_{t+1} \sim \mathcal{N}(\mu_{t+1}, \log \sigma^2_{t+1})$ for performing Slot Attention with visual features at frame $t + 1$.

To enforce this stochasticity, we apply a KL divergence loss on the predicted distribution.
Since we do not regularize the mean of $\tilde{\mathcal{S}}_{t+1}$, the loss only penalizes the log variance with a prior value $\hat{\sigma}$:
\begin{align}
\begin{split}
    \mathcal{L}_{KL}^{t+1} &= D_{\mathrm{KL}}(\mathcal{N}(\mu_{t+1}, \log \sigma^2_{t+1})\ ||\ \mathcal{N}(\mu_{t+1}, \log \hat{\sigma}^2))\\
    &= \log \frac{\hat{\sigma}}{\sigma_{t+1}} + \frac{\sigma_{t+1}^2}{2 \cdot \hat{\sigma}^2} - \frac{1}{2},
\end{split}
\end{align}
which will be averaged over all input timesteps.
We set $\hat{\sigma} = 0.1$ which produces enough randomness to break the symmetry without destroying the temporal alignment of slots.
With this simple modification, we can detect all the objects throughout the video.
We use the same strategy as SAVi to train the stochastic SAVi model on CLEVRER under a combination of the frame reconstruction loss and the KL divergence loss, where the later one is weighted by a factor of $1 \times 10^{-4}$.

\heading{STEVE.}
We reproduce the $128 \times 128$ input resolution version of STEVE.
Different from the paper, we adopt a two-stage training strategy by first pre-training a discrete VAE~\citep{SLATE} to convert images into patches tokens, and then train the slot model to reconstruction these tokens.
We found this strategy lead to more stable training in our experiments.
Other training settings are the same as their original implementation.

\heading{Transformer.}
We follow BERT~\citep{BERT} to implement our model by stacking multiple transformer encoder blocks.
The number of self-attention head is 8 and the hidden size of FFN is $4 \times D_e$.
We adopt the Pre-LN Transformer~\citep{Pre-LN-Transformer} design as we empirically find it easier to optimize.
We train our model using the Adam optimizer.
The initial learning rate is $2 \times 10^{-4}$ and decayed to $0$ in a cosine schedule.
We also adopt a linear learning rate warmup strategy during the first $5\%$ of training steps.
We do not apply gradient clipping or weight decay during training.
On OBJ3D and CLEVRER, we apply both the slot reconstruction loss $\mathcal{L}_S$ and image reconstruction loss $\mathcal{L}_I$ for training.
On Physion, we do not apply $\mathcal{L}_I$ due to the large memory consumption of STEVE's Transformer-based decoder.
Similarly, we do not apply $\mathcal{L}_I$ on PHYRE since the image resolution is $128 \times 128$ and the spatial broadcast size of the SAVi decoder is set as $16 \times 16$, which consumes lots of GPU memory.

\heading{VQA model on CLEVRER.}\label{appendix:aloe}
To jointly process object slots and question texts, we employ Aloe~\citep{Aloe} as the base VQA model given its strong performance on CLEVRER.
Given a video and a question, Aloe first leverages pre-trained object-centric model to extract slots from each frame, and a text tokenizer to convert questions to language tokens.
Then, it concatenates slots and text tokens, and forward them to a reasoning module, which is a stack of $N_{Aloe}$ Transformer encoder, to perform joint reasoning and predict the answer.

We re-implement Aloe in PyTorch.
Following their training settings, we reproduce the results with a smaller Transformer reasoning module, $N_{Aloe}$ = $12$, while the original implementation uses $N_{Aloe}$ = $28$.
This is because we use SAVi which produces higher quality and temporally consistent slots compared to the MO-Net~\citep{MONet} they used.
When integrating with \algoNameFull, we unroll our dynamics model to predict slots at future timesteps, and feed them to Aloe to answer predictive questions.
For other types of questions, the process is the same as the original Aloe.

\heading{Readout model on Physion.}\label{appendix:physion-readout}
Permutation equivariance is an important property of object-centric models, which is also preserved by \algoNameFull.
Simply concatenating slots and forwarding it through a fully-connected (FC) layer degrades the performance, since the prediction changes according to input slot orders.
To build a compatible readout model, we leverage the max-pooling operation which is invariant to the input permutations~\citep{PointNet}.
Besides, to better utilize the object-centric representation, we draw inspiration from Relation Networks~\citep{RelationNetwork} to explicitly reason over pairs of objects.
Specifically, we concatenate every two slots and apply FC on it, then the outputs are max-pooled over all pairs of slots and time to obtain the final prediction.

In our experiments, we discovered that training readout models on the entire rollout videos (150 frames) leads to severe overfitting.
This is because of the error accumulation issue in long-term video prediction, where the model overfits to artifacts introduced at later timesteps.
Therefore, we only fit the readout network to the first 75 frames of the video.
We evaluate baselines with the same readout model and training setting for fair comparison, which also improves their performance.

\heading{Task success classifier on PHYRE.}\label{appendix:phyre-readout}
We train a task success classifier to score an action for the current task.
Specifically, we concatenate predicted slots with a learnable \texttt{CLS} token, add temporal positional encoding, and process them using a Transformer encoder.
Then, we apply a two-layer MLP to output the score from the features corresponding to the \texttt{CLS} token.
Such design also ensures the predicted score is invariant to the order of input slots.


\section{Baselines}\label{appendix:baselines}

We detail our implementation of baselines in this section.


\heading{PredRNN~\citep{PredRNN}} is a famous video prediction model leveraging spatio-temporal LSTM to model scene dynamics via global frame-level features.
We adopt the online official implementation \footnote{\href{https://github.com/thuml/predrnn-pytorch}{https://github.com/thuml/predrnn-pytorch}}.
The models are trained until convergence for 16 epochs and 6 epochs on OBJ3D and CLEVRER, respectively.
We adopt the same training settings as their original paper.

\heading{VQFormer.}
To show the effectiveness of object-centric representations, we design a baseline that replace the object slots with Vector Quantized (VQ) patch tokens.
We first pre-train a VQ-VAE~\citep{VQVAE} on video frames to convert patches to discrete tokens.
We adopt the implementation of VQ-VAE from VQ-GAN~\citep{VQGAN} \footnote{\href{https://github.com/CompVis/taming-transformers}{https://github.com/CompVis/taming-transformers}}, where we set the number of tokens per-frame as $4 \times 4 = 16$, and the codebook size as 4096.
The autoregressive Transformer follows the design in \algoNameFull.
We also tried the GPT-like training strategy (i.e. causal masking) as done in \citet{TransformerWorldModel}, but did not observe improved performance.

\heading{G-SWM~\citep{G-SWM}} unifies several priors in previous object-centric models and is shown to achieve good results on various simple video datasets.
It constructs a background module to process the scene context, disentangles object features to positional and semantic information, explicitly models occlusion and interaction using depth and GNN module, and performs hierarchical latent modeling to deal with the multi-modality over time.
We use the online official implementation \footnote{\href{https://github.com/zhixuan-lin/G-SWM}{https://github.com/zhixuan-lin/G-SWM}}.
We train the model for 1M steps on both datasets, and select the best weight via the loss on the validation set.
Our re-trained model achieves slightly better results than their pretrained weight on the OBJ3D dataset.
Therefore, we also adopt the this training setting on CLEVRER.

\heading{SAVi-dyn.}
Since neither the code of PARTS~\citep{PARTS} nor its testing dataset (PLAYROOM) is released, and PARTS did not try future prediction task on CLEVRER, we try our best to re-implement it to compare with \algoNameFull under our settings.
Inspired by its design, we replace the Transformer predictor in SAVi~\citep{SAVi} with the Transformer-LSTM dynamics module in PARTS.
The model is trained to observe initial burn-in frames, and then predict the slots as well as the reconstructed image of the rollout frames using the dynamics module.
We use a learning rate of $1 \times 10^{-4}$ and train the model for 500k steps.
The other training strategies follow SAVi.

We do not compare with OCVT~\citep{OCVT} because it underperforms G-SWM even on simple 2D datasets, while \algoNameFull outperforms G-SWM under all the settings.

\heading{Aloe~\citep{Aloe}} runs Transformers over object slots and text tokens of the question to perform reasoning.
The official code \footnote{\href{https://github.com/deepmind/deepmind-research/tree/master/object\_attention\_for\_reasoning}{https://github.com/deepmind/deepmind-research/tree/master/object\_attention\_for\_reasoning}} was written in TensorFlow~\citep{TensorFlow}, so we re-implement it in PyTorch to fit in our codebase.
We adopt the same model architecture and hyper-parameters as the original paper, except that we use 12 layer Transformer encoder while they use 28, as our SAVi slot representations are more powerful than their MO-Net~\citep{MONet} slots.
We train the model for 250k steps.

\heading{RPIN~\citep{RPIN}} is an object-centric dynamics model using ground-truth object bounding boxes of the burn-in frames.
For a fair comparison on Physion dataset with \algoNameFull, we also apply our improved readout model (see Appendix~\ref{appendix:physion-readout}) on RPIN.
We adopt the online official implementation \footnote{\href{https://github.com/HaozhiQi/RPIN}{https://github.com/HaozhiQi/RPIN}} and train it on Physion dataset for 300k steps.
As shown in Table~\ref{table:physion-vqa-numbers}, our reproduced readout accuracy is much higher than the reported result in the benchmark.

\heading{pDEIT-lstm} applies an LSTM over frame features extracted by ImageNet~\citep{ImageNet} pre-trained DEIT~\citep{DEIT} model.
We follow the original implementation and use the DEIT model provided by timm~\citep{TIMM} (\texttt{deit\_base\_patch16\_224}).
We frozen the DEIT model and train the LSTM for 100k steps.

For other baselines, we simply copy the numbers from previous papers.

\begin{figure*}[t]
    \vspace{\pagetopmargin}
    \vspace{-2mm}
    \centering
    \subfigure{
        \includegraphics[width=0.98\textwidth]{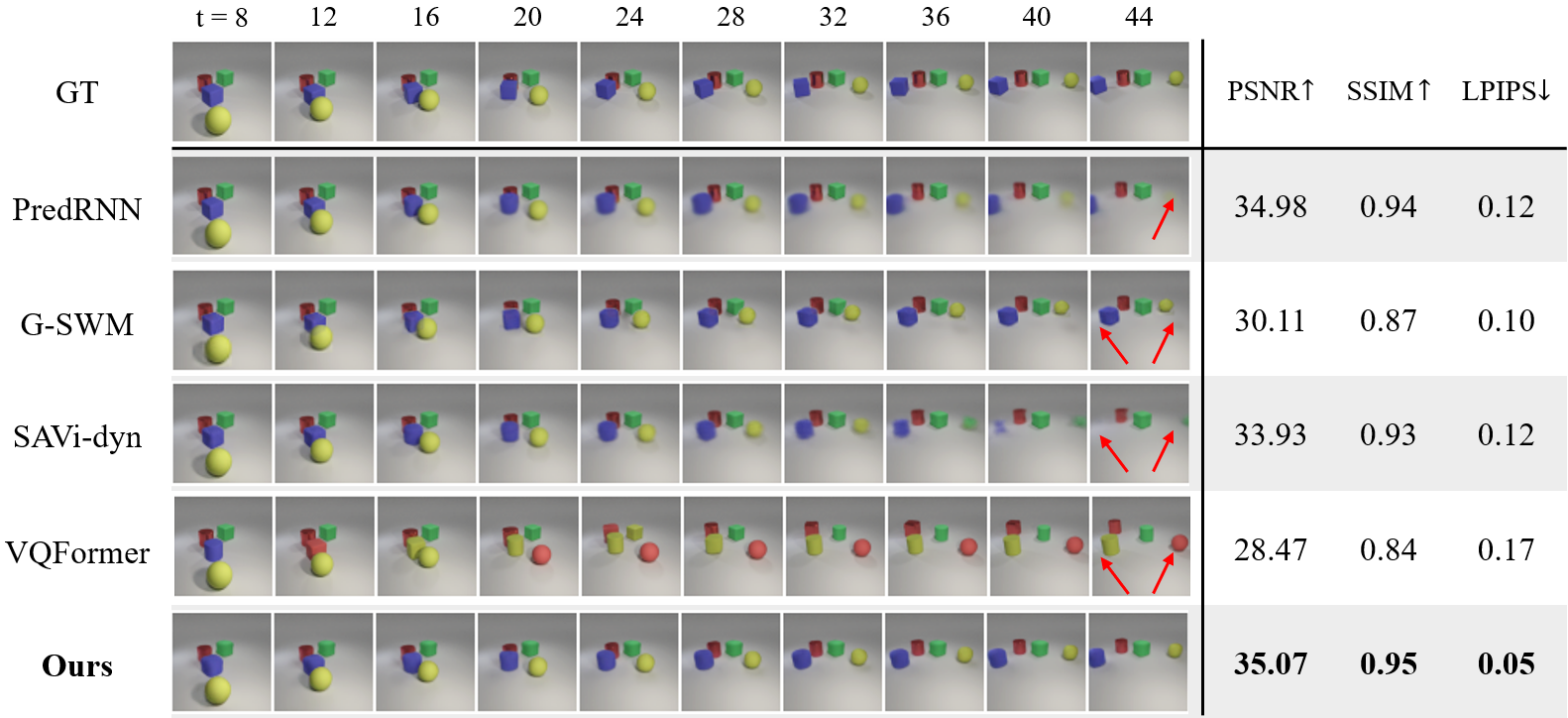}
    }\\
    \vspace{-3mm}
    \subfigure{
        \includegraphics[width=0.98\textwidth]{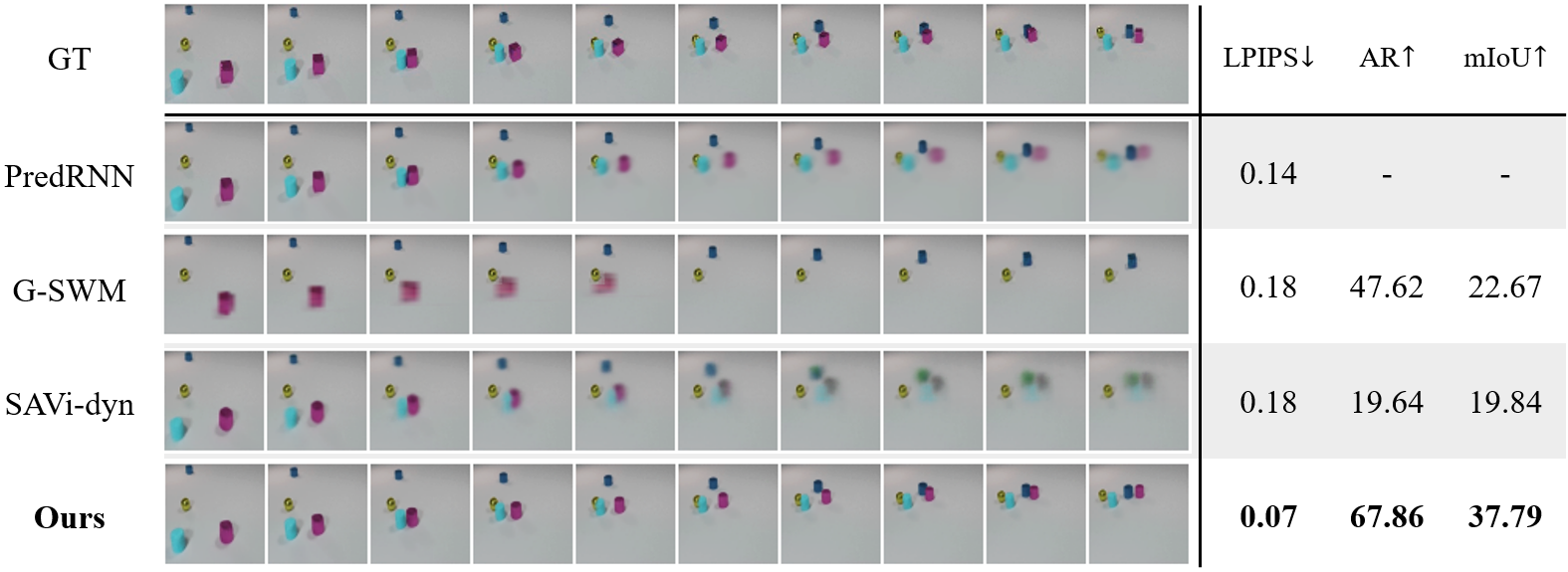}
    }\\
    \vspace{\figcapmargin}
    \caption{
    Generation results on OBJ3D (top) and CLEVRER (bottom).
    On the right, we report metrics measuring the visual quality and object trajectory of the visualized rollouts for each model.
    }
    \label{appendix-fig:obj3d-clevrer-vis}
    \vspace{\figmargin}
\end{figure*}

\begin{figure*}[!t]
    \vspace{-0.5mm}
    \centering
    \includegraphics[width=0.99\linewidth]{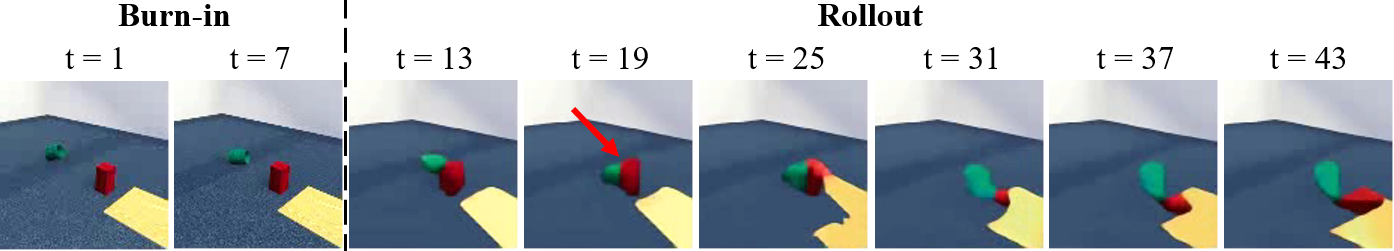}
    \caption{
    Qualitative results on Physion VQA task.
    To answer the question ``\textit{Will the red object contact with the yellow object?}", \algoNameFull successfully simulates the falling of the red box.
    The low visual quality of the predicted frames is due to STEVE's Transformer-based decoder, which is not designed for pixel-space reconstruction.
    Nevertheless, they still preserve the correct motion of objects.
    }
    \label{appendix-fig:vqa-vis}
    \vspace{-1.5mm}
\end{figure*}

\section{More Experimental Results}

\subsection{Qualitative Results}\label{appendix:more-vis}

\heading{Video prediction.}
Figure~\ref{appendix-fig:obj3d-clevrer-vis} (top) shows additional qualitative results on OBJ3D.
\algoNameFull achieves excellent generation of the object trajectories thus very low LPIPS score.
However, its PSNR and SSIM are still close to PredRNN and SAVi-dyn, which blurs the moving objects into the background in later frames.
This again proves that LPIPS are superior metrics for measuring the generated videos.
Besides, G-SWM can also preserve the object identity because it leverages complex priors such as depth to model occlusions.
Nevertheless, its simulated dynamics are still worse than our Transformer model.
\revised{
Finally, VQFormer is able to generate sharp images without blurry objects because of its strong VQ-VAE decoder.
However, the object properties such as colors are not consistent, and the simulated dynamics are erroneous.
}

We present a visual result on CLEVRER in Figure~\ref{appendix-fig:obj3d-clevrer-vis} (bottom).
The objects are smaller in size and have longer term dynamics, making it much harder than OBJ3D.
PredRNN and SAVi-dyn still generate blurry objects at later steps.
G-SWM sometimes cannot detect objects newly entering the scene because of the limited capacity of its discovery module.
In contrast, \algoNameFull builds Transformer on SAVi slots, enabling both accurate object detection and precise dynamics modeling.
This is also verified by the object-aware metrics AR and mIoU we show in the figures.

\begin{figure*}[!hb]
    \vspace{6mm}
    \centering
    \subfigure{
        \includegraphics[width=0.98\textwidth]{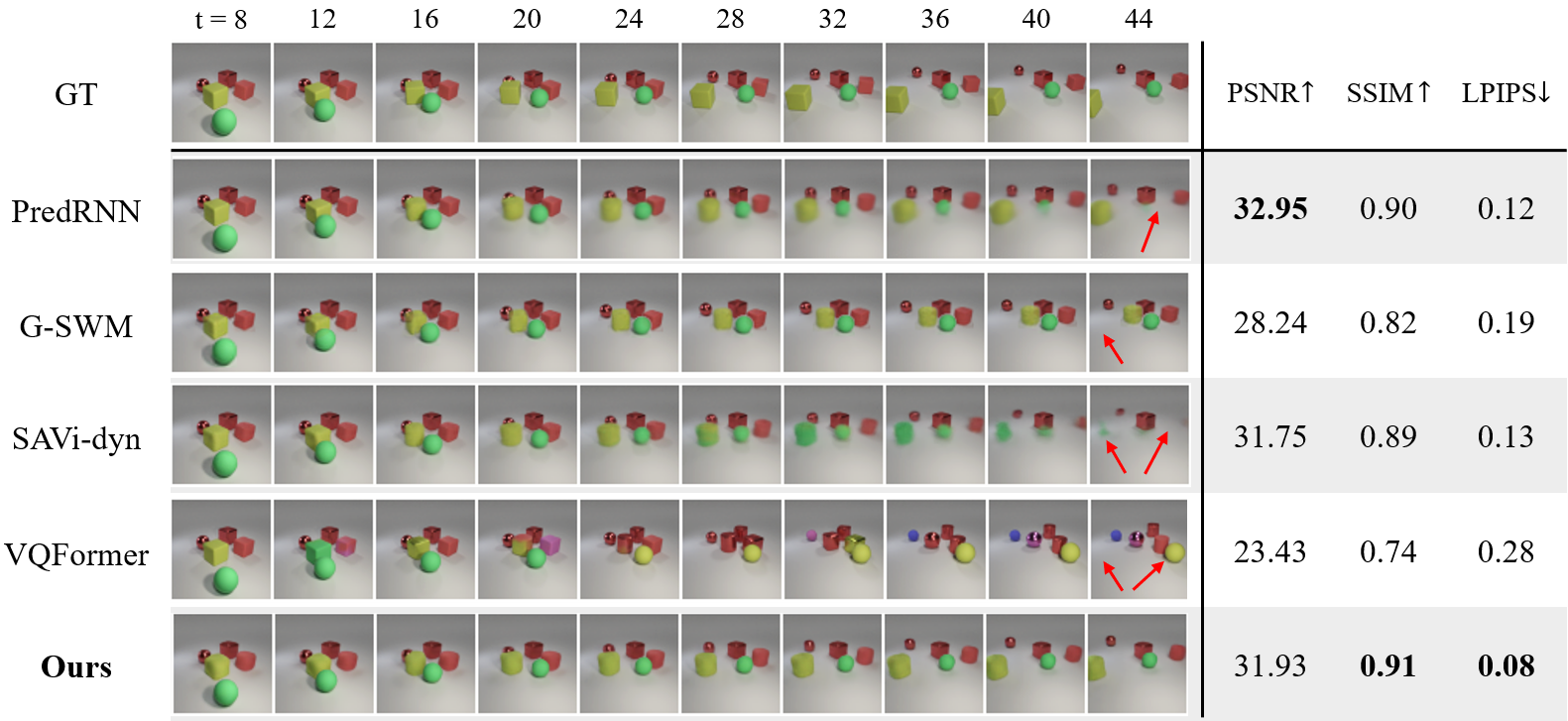}
    }\\
    \vspace{-2mm}
    \subfigure{
        \includegraphics[width=0.98\textwidth]{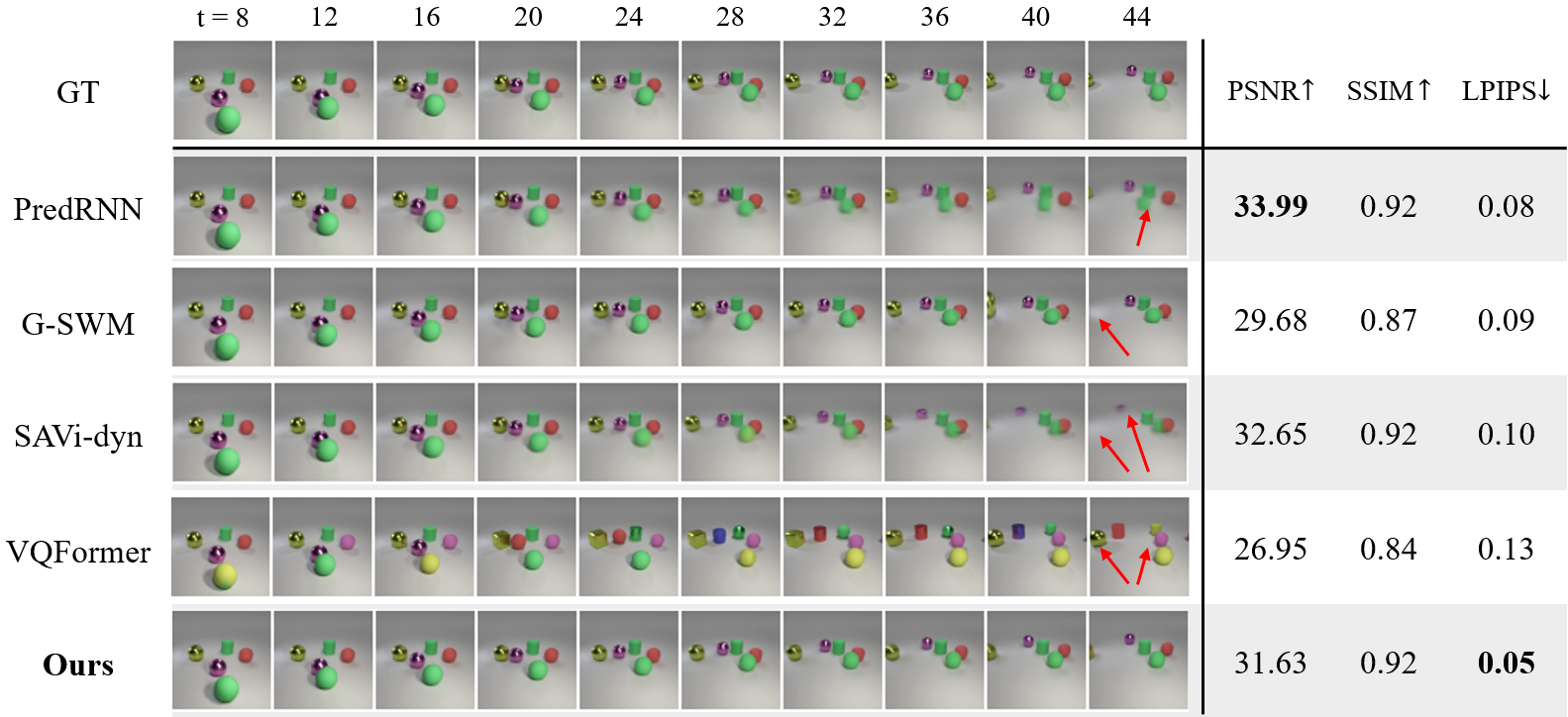}
    }\\
    \caption{
    More qualitative results on OBJ3D.
    }
    \label{appendix-fig:more-obj3d-vis}
\end{figure*}

\begin{figure*}[t]
    \centering
    \subfigure{
        \includegraphics[width=0.98\textwidth]{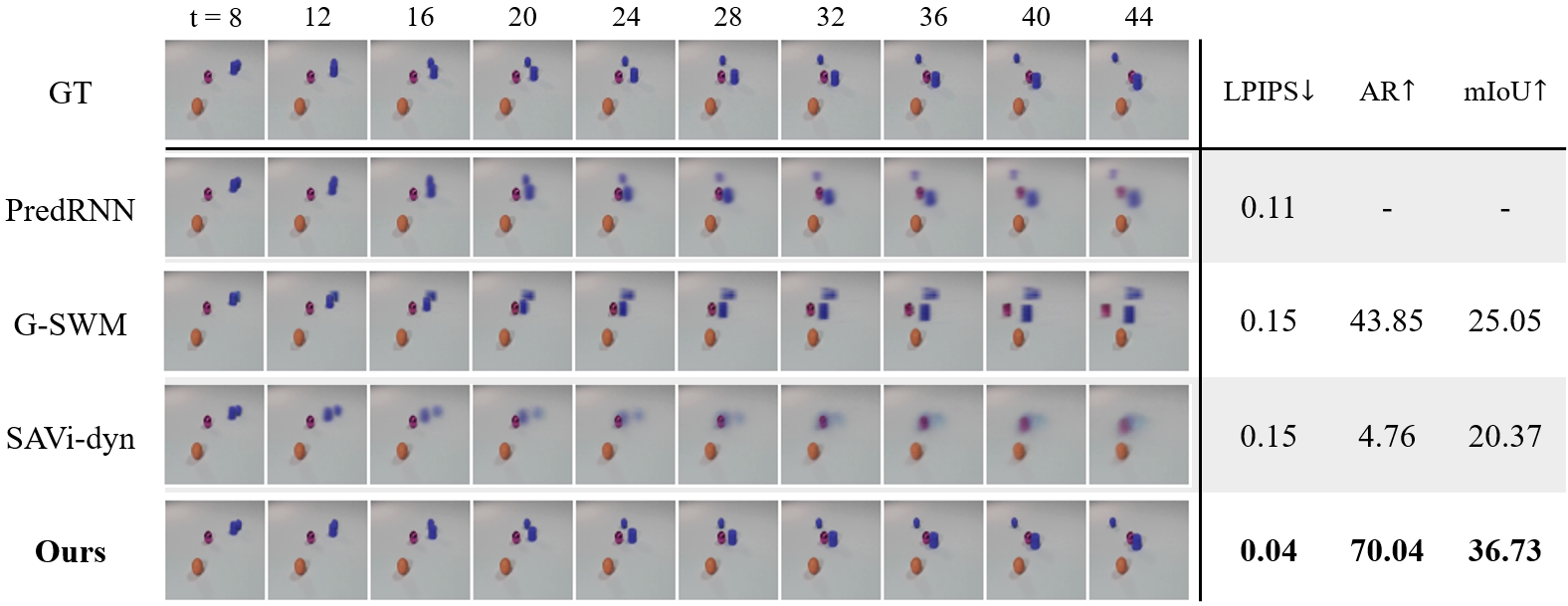}
    }\\
    \vspace{-2mm}
    \subfigure{
        \includegraphics[width=0.98\textwidth]{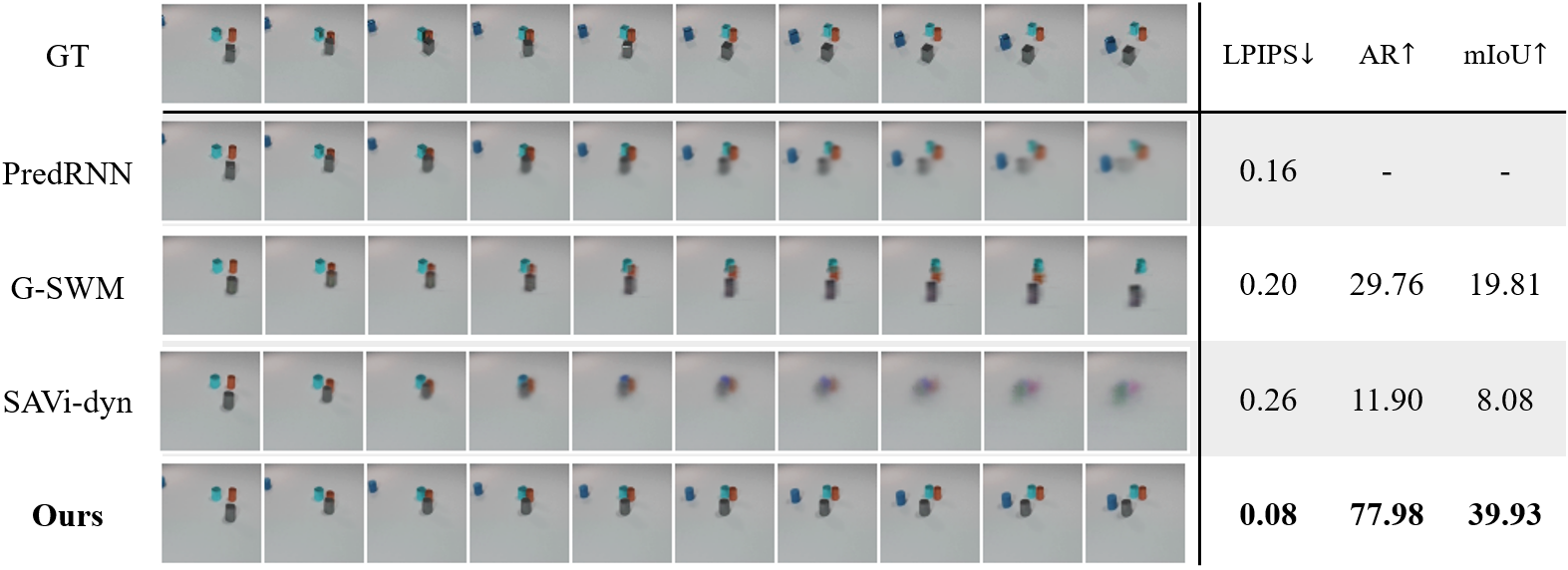}
    }\\
    \caption{
    More qualitative results on CLEVRER.
    }
    \label{appendix-fig:more-clevrer-vis}
\end{figure*}

See Figure~\ref{appendix-fig:more-obj3d-vis} and Figure~\ref{appendix-fig:more-clevrer-vis} for more qualitative results in the video prediction task.

\heading{VQA.}
Figure~\ref{appendix-fig:vqa-vis} shows a qualitative result on Physion dataset, where \algoNameFull successfully synthesizes the contact of the red object and the yellow ground.
Note the low quality of the predicted frames is due to the STEVE's Transformer-based decoder, which is not designed for pixel-space reconstruction \footnote{In our experiments, even the reconstruction results of STEVE on Physion videos are of low quality.}.
Improving the decoder design is beyond the scope of this paper.

\begin{figure*}[t]
    \vspace{\pagetopmargin}
    \centering
    \subfigure{
        \includegraphics[width=0.49\textwidth]{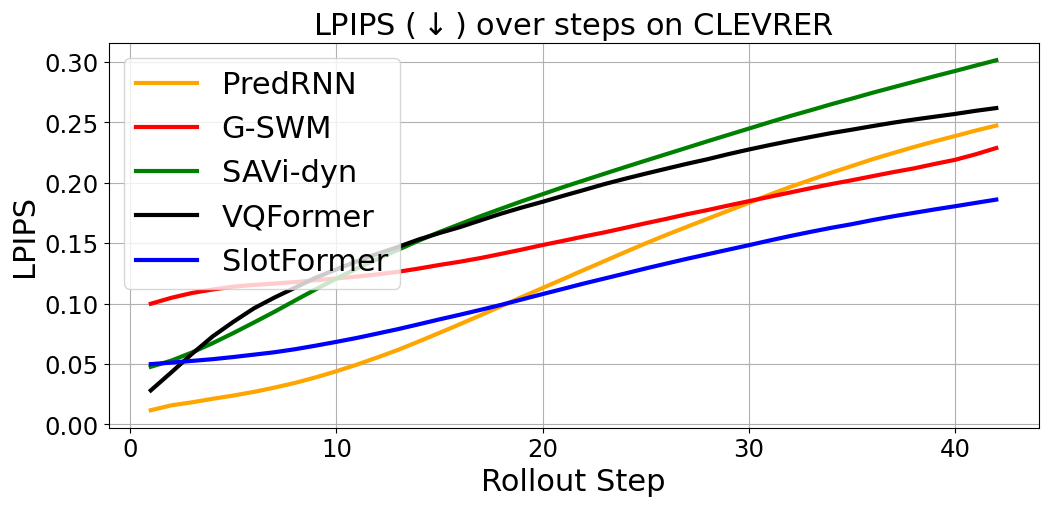}
    }
    \hspace{-3mm}
    \subfigure{
        \includegraphics[width=0.49\textwidth]{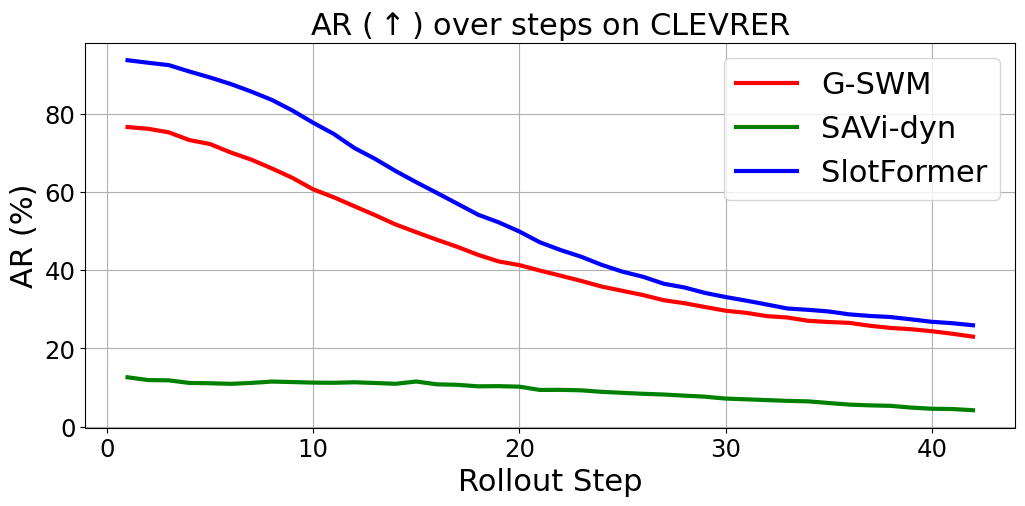}
    }
    \\
    \vspace{-3mm}
        \subfigure{
        \includegraphics[width=0.49\textwidth]{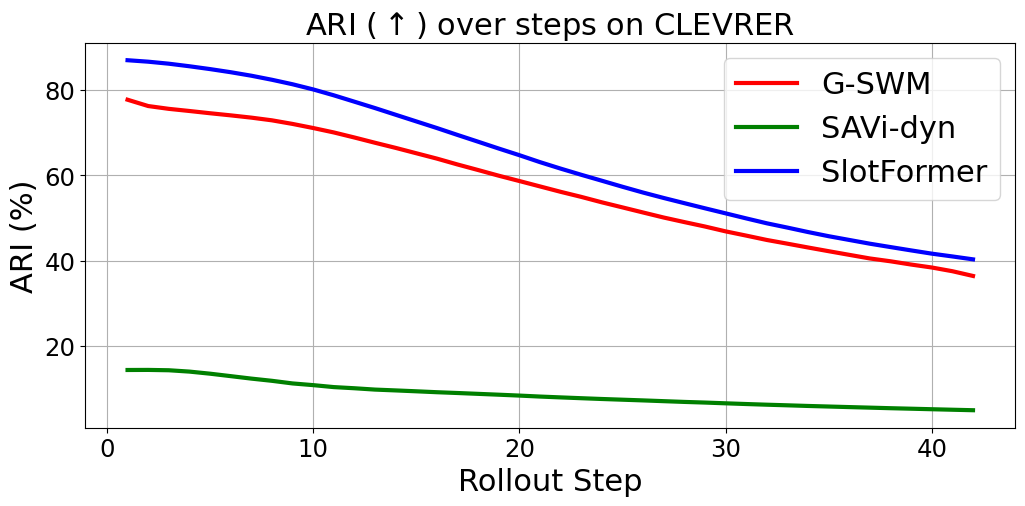}
    }
    \hspace{-3mm}
    \subfigure{
        \includegraphics[width=0.49\textwidth]{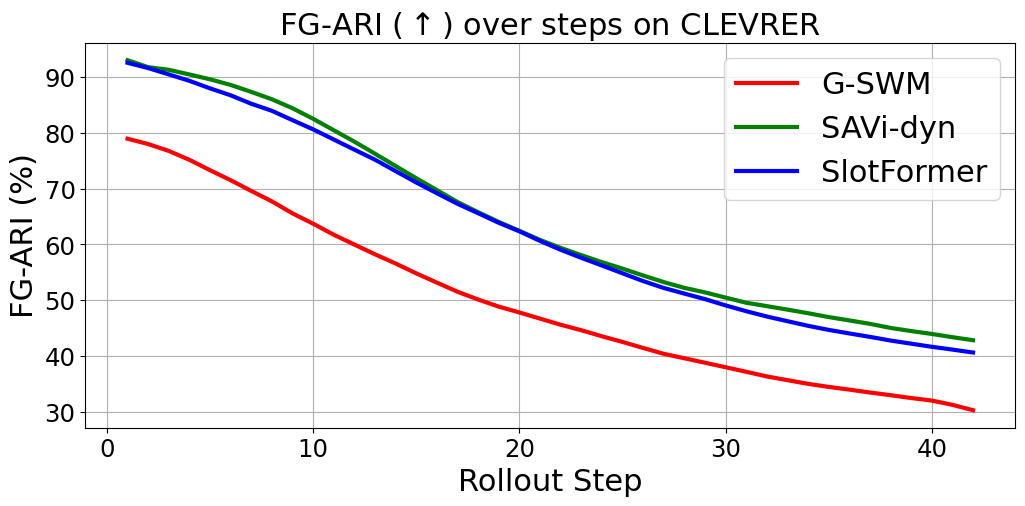}
    }
    \vspace{\figcapmargin}
    \caption{
    Comparison of the object dynamics of the generated videos at each rollout step on CLEVRER.
    We report FG-API (left) and FG-mIoU (right) of the segmentation masks.
    }
    \label{appendix-fig:object-dynamics-step-lines}
\end{figure*}

\begin{table}[t]
    \vspace{\pagetopmargin}
    \begin{minipage}{.48\linewidth}
        \centering
        \small
        \setlength{\tabcolsep}{3pt}
        \begin{tabular}{lcc}
        \toprule
        \multirow{2}{*}{\textbf{Method}} & \multicolumn{2}{c}{\textbf{Predictive}} \\
        & per opt. ($\%$) & per ques. ($\%$) \\
        \midrule
        DCL & 90.5 & 82.0 \\
        VRDP & 91.7 & 83.8 \\
        VRDP$^\dag$ & 94.5 & 89.2 \\
        Aloe & 93.5 & 87.5 \\
        Aloe$^*$ & 93.1 & 87.3 \\
        \midrule
        Aloe$^*$ + \textbf{Ours} & \textbf{96.5} & \textbf{93.3} \\
        \bottomrule
    \end{tabular}
    \caption{
    Predictive VQA on CLEVRER, reporting per-option (per opt.) and per-question (per ques.) accuracy.
    DCL and VRDP$^\dag$ both utilize pre-trained object detectors;
    $^*$ indicates our re-implementation.
    \label{appendix-table:full-clevrer-vqa-numbers}
    }
    \end{minipage}%
    \hfill
    \begin{minipage}{.48\linewidth}
        \centering
        \small
        \setlength{\tabcolsep}{4pt}
        \begin{tabular}{lcc|c}
        \toprule
        \textbf{Method} & Obs. ($\%$) & Dyn. ($\%$) & $\uparrow$ ($\%$) \\
        \midrule
        \textcolor{gray}{Human} & \textcolor{gray}{74.7} & \textcolor{gray}{-} & \textcolor{gray}{-} \\
        \midrule
        RPIN & 54.3 & 55.7 & + 1.4 \\
        RPIN$^*$ & 62.8 & 63.8 & + 1.0 \\
        pDEIT-lstm & 59.9 & 60.5 & + 0.6 \\
        pDEIT-lstm$^*$ & 59.2 & 60.0 & + 0.8 \\
        \midrule
        \textbf{Ours} & \textbf{65.2} & \textbf{67.1} & \textbf{+ 1.9} \\
        \bottomrule
    \end{tabular}
    \caption{
    VQA accuracy on Physion.
    We report the readout accuracy on observation (OBS.) and observation plus rollout (Dyn.) frames.
    $\uparrow$ denotes the improvement brought by the learned dynamics.
    Methods marked with $^*$ are our reproduced results.
    \label{appendix-table:full-physion-vqa-numbers}
    }
    \end{minipage}
\end{table}

\begin{table}[t]
    \vspace{\pagetopmargin}
    \centering
    \small
    \resizebox{1.0\textwidth}{!}{
    \setlength{\tabcolsep}{3pt}
    \begin{tabular}{lcccccccccc}
        \toprule
        \multirow{2}{*}{\textbf{Method}} & \multirow{2}{*}{\textbf{Descriptive}} & \multicolumn{2}{c}{\textbf{Explanatory}} & \multicolumn{2}{c}{\textbf{Predictive}} & \multicolumn{2}{c}{\textbf{Counterfactual}} & \multirow{2}{*}{\textbf{Average}} \\
        & & per opt. & per ques. & per opt. & per ques. & per opt. & per ques. & \\
        \midrule
        Aloe$^*$ + \textbf{Ours} & 95.17 & 98.04 & 94.79 & 96.50 & 93.29 & 90.63 & 73.78 & 89.26 \\
        \bottomrule
    \end{tabular}
    }
    \caption{
    Accuracy on different questions and average results on CLEVRER.
    Numbers are in $\%$.
    \label{appendix-table:clevrer-vqa}
    }
    \vspace{-3mm}
\end{table}

\begin{table}[!t]
    \centering
    \small
    \setlength{\tabcolsep}{3pt}
    \begin{tabular}{lccccccccc}
        \toprule
        \textbf{Method} & \textbf{Collide} & \textbf{Contain} & \textbf{Dominoes} & \textbf{Drape} & \textbf{Drop} & \textbf{Link} & \textbf{Roll} & \textbf{Support} & \textbf{Average} \\
        \midrule
        \algoNameFull & 69.3 & 63.3 & 55.6 & 66.7 & 62.7 & 69.3 & 70.7 & 77.3 & 67.1 \\
        \bottomrule
    \end{tabular}
    \caption{
    Accuracy breakdown for all eight scenarios on Physion.
    Numbers are in $\%$.
    \label{appendix-table:physion-vqa}
    }
    \vspace{-3mm}
\end{table}

\begin{table}[!t]
    \centering
    \small
    \begin{tabular}{lccccccccccc}
        \toprule
        \multirow{2}{*}{\textbf{Method}} & \multicolumn{10}{c}{\textbf{Fold ID}} & \multirow{2}{*}{\textbf{Average}} \\
        & 0 & 1 & 2 & 3 & 4 & 5 & 6 & 7 & 8 & 9 & \\
        \midrule
        \algoNameFull & 83.1 & 83.2 & 81.0 & 81.2 & 81.2 & 83.0 & 82.6 & 80.0 & 83.0 & 81.8 & \pms{82.0}{1.1} \\
        \bottomrule
    \end{tabular}
    \caption{
    AUCCESS for all 10 folds on PHYRE.
    \label{appendix-table:phyre-fold-breakdown}
    }
    \vspace{-3mm}
\end{table}

\begin{table}[t!]
    \begin{minipage}{.55\linewidth}
    \small
    \resizebox{0.99\textwidth}{!}{
    \setlength{\tabcolsep}{2pt}
    \begin{tabular}{lccc|ccc}
        \toprule
        \multirow{2}{*}{\textbf{Method}} & \multicolumn{3}{c}{\textbf{OBJ3D}} & \multicolumn{3}{c}{\textbf{CLEVRER}} \\
        & PSNR $\uparrow$ & SSIM $\uparrow$ & LPIPS $\downarrow$ & PSNR $\uparrow$ & SSIM $\uparrow$ & LPIPS $\downarrow$ \\
        \midrule
        OCVT & 31.08 & 0.88 & 0.13 & 27.96 & 0.87 & 0.18 \\
        Slot-LSTM & 32.15 & 0.90 & 0.09 & 29.79 & 0.88 & 0.13 \\
        \midrule
        \textbf{Ours} & \textbf{32.40} & \textbf{0.91} & \textbf{0.08} & \textbf{30.21} & \textbf{0.89} & \textbf{0.11} \\
        \bottomrule
    \end{tabular}
    }
    \caption{
    \revised{
    Evaluation of visual quality on both datasets.
    }
    \label{appendix-table:vp-visual-quality}
    }
    \end{minipage}%
    \hfill
    \begin{minipage}{.445\linewidth}
    \small
    \resizebox{0.99\textwidth}{!}{
    \setlength{\tabcolsep}{2pt}
    \begin{tabular}{lcccc}
        \toprule
        \textbf{Method} & AR $\uparrow$ & ARI $\uparrow$ & FG-ARI $\uparrow$ & FG-mIoU $\uparrow$ \\
        \midrule
        OCVT & 36.19 & 51.23 & 40.87 & 20.57 \\
        Slot-LSTM & 48.52 & 59.58 & 58.42 & 27.84 \\
        \midrule
        \textbf{Ours} & \textbf{53.14} & \textbf{63.45} & \textbf{63.00} & \textbf{29.81} \\
        \bottomrule
    \end{tabular}
    }
    \caption{
    \revised{
    Evaluation of object dynamics on CLEVRER.
    All the numbers are in $\%$.
    }
    \label{appendix-table:vp-object-dynamics}
    }
    \end{minipage}
\end{table}

\subsection{Quantitative Results}

\heading{Video prediction.}\label{appendix:per-step-dynamics}
We show the per-step FG-ARI and FG-mIoU results in Figure~\ref{appendix-fig:object-dynamics-step-lines}.
The sophisticated priors in G-SWM prevents it from scaling to scenes with multiple objects and complex dynamics.
Since SAVi-dyn generates blurry objects, it produces many false positives in the segmentation masks.
Instead, \algoNameFull preserves the object identity and achieves high scores in both metrics over long rollout steps.

\heading{VQA.}
Table~\ref{appendix-table:full-clevrer-vqa-numbers} and Table~\ref{appendix-table:full-physion-vqa-numbers} present the complete results of the original and our reproduced performance of baselines, as well as ours on both VQA datasets.
We report the performance of our Aloe with \algoNameFull model on all four question types of CLEVRER in Table~\ref{appendix-table:clevrer-vqa}.
We report the per-scenario accuracy of \algoNameFull on Physion rollout setting in Table~\ref{appendix-table:physion-vqa}.

\heading{Planning.}
Table~\ref{appendix-table:phyre-fold-breakdown} shows the AUCCESS of \algoNameFull for all 10 folds on PHYRE.

\subsection{\revised{Comparison with Additional Baselines}}\label{appendix:more-baseline}

\revised{
In this section, we compare \algoNameFull with two additional baselines, namely \textit{OCVT}~\citep{OCVT} and \textit{Slot-LSTM}, in the video prediction task on OBJ3D and CLEVRER datasets.
}

\revised{
\heading{OCVT}
builds Transformers over SPACE~\citep{SPACE} which applies heavy priors in their framework.
In addition, it requires a Hungarian alignment step of slots for loss computation.
Therefore, OCVT underperforms G-SWM~\citep{G-SWM} in long-term generation.
Since its code is not released, we reproduce it based on SPACE \footnote{\href{https://github.com/zhixuan-lin/SPACE}{https://github.com/zhixuan-lin/SPACE}}, and adopt its setting in the video prediction task.
}

\revised{
As shown in Table~\ref{appendix-table:vp-visual-quality} and Table~\ref{appendix-table:vp-object-dynamics}, OCVT underperforms \algoNameFull in both visual quality of videos and accuracy of object dynamics.
This is because object slots from SAVi is more powerful than SPACE, and \algoNameFull naturally enjoys the temporal alignment of slots.
}

\revised{
\heading{Slot-LSTM}
trains a Transformer-LSTM dynamics module from PARTS~\citep{PARTS} over the same pre-trained object slots as \algoNameFull.
We adopt the same Transformer module as \algoNameFull, but only feed in slots at a single timestep, thus only modeling the spatial interaction of objects.
The Transformer is followed by a per-slot LSTM to learn the temporal dynamics.
}

\revised{
As shown in Table~\ref{appendix-table:vp-visual-quality} and Table~\ref{appendix-table:vp-object-dynamics}, \algoNameFull outperforms Slot-LSTM in all the metrics, especially on the more challenging CLEVRER dataset.
This indicates the importance of joint spatial-temporal reasoning over a larger context window.
Despite having the same Transformer module, Slot-LSTM limits the context window of its recurrent module to only a single timestep.
Therefore, it still generates videos with blurry objects and inconsistent dynamics over the long horizon.
}

\begin{figure*}[t]
    \vspace{\pagetopmargin}
    \vspace{-2mm}
    \centering
    \subfigure{
        \includegraphics[width=0.49\textwidth]{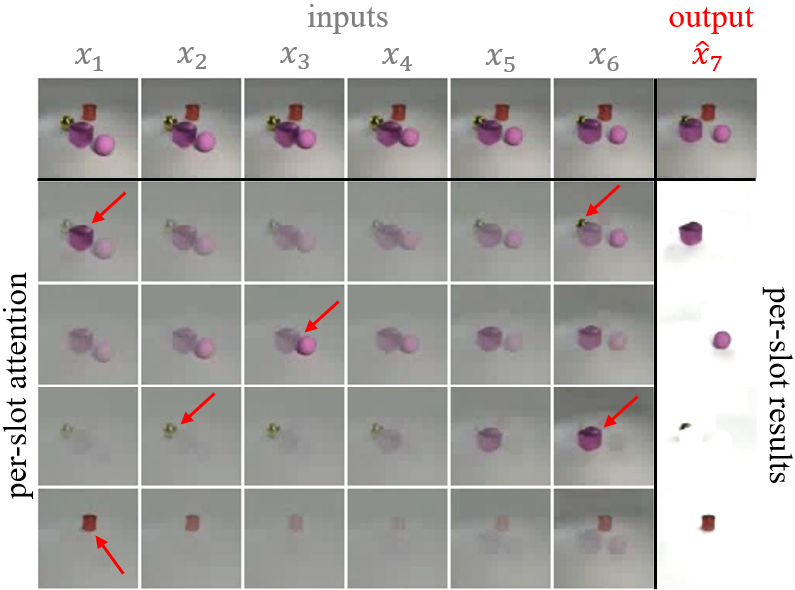}
    }
    \hspace{-3mm}
    \subfigure{
        \includegraphics[width=0.49\textwidth]{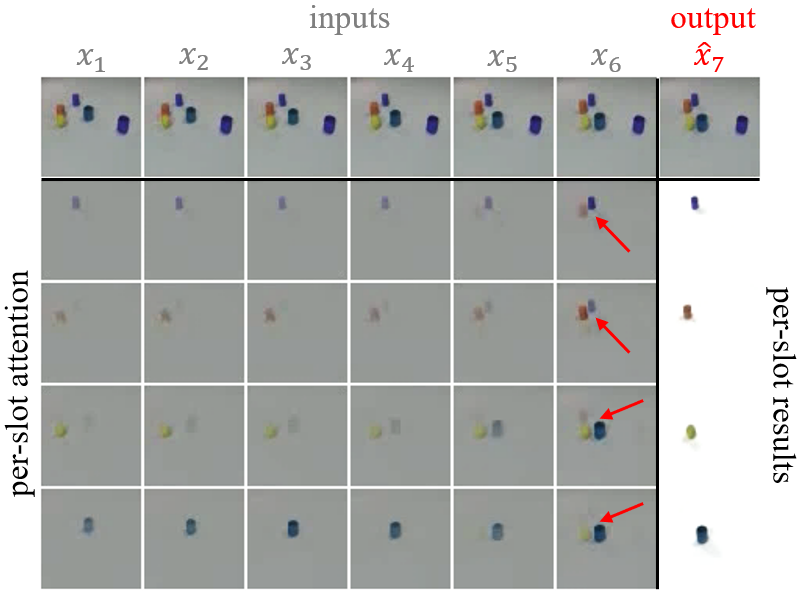}
    }
    \vspace{-2mm}
    \caption{
    Example attention map visualization on OBJ3D (left) and CLEVRER (right).
    Our model takes in slots from $\{\bm{x}_i\}_{i=1}^6$ (column 1-6) to predict slots of $\hat{\bm{x}}_7$ (column 7).
    We show images at the first row and the per-slot future reconstructions at the rightmost column.
    The body of the table shows the per-slot attention of \algoNameFull when predicting $\hat{\mathcal{S}}_7$, with the arrows pointing at the regions of high importance for predicting the future slot in the same row.
    Zoom in for better viewing.
    }
    \label{appendix-fig:attn-map}
    \vspace{-1mm}
\end{figure*}

\subsection{Attention Analysis}\label{appendix:exp-attention-map}

In this section, we analyze the visual cues in the input frames that \algoNameFull utilizes to make future predictions.
We do so by visualizing the attention map from the last self-attention layer in the transformer $\mathcal{T}$.
More precisely, given the last $T$ encoded frames $\{\mathcal{S}_t\}_{t=1}^T$, we are predicting the future slots $\hat{\mathcal{S}}_{T+1}$.
Denote the attention scores from $\hat{\bm{s}}_{T+1}^i$ to $\bm{s}_t^j$ as $\bm{a}^i_{t, j}$, where $i, j \in [1, N]$ and $N$ is the number of slots.
At each timestep $t$ and for each future slot $i$, we obtain spatial attention maps $o^i_t$ over input frames $x_i$ as a weighted combination of the slot reconstructions as follows:
\begin{equation}
    \bm{o}^i_t = \sum_{j=1}^N \bm{a}^i_{t, j} \cdot (\bm{m}_t^j \odot \bm{y}_t^j),
\end{equation}
which indicates the regions of $\bm{x}_t$ \algoNameFull attends upon when predicting $\hat{s}_{T+1}^i$.

Figure~\ref{appendix-fig:attn-map} (left) presents one example from OBJ3D, where the purple cube just collided with the purple sphere, and is about to hit the yellow sphere.
When predicting the purple cube, the model focuses on the past collision event in $\{\bm{x}_i\}_{i=1}^4$, and highlights the yellow sphere in $\bm{x}_6$.
For the purple sphere, the Transformer only looks at the purple cube because it will not hit the yellow sphere.
Since the yellow sphere becomes heavily occluded in $\bm{x}_6$, \algoNameFull attends to earlier frames, while predicting its future motion based on the purple cube.
\revised{
This indicates that \algoNameFull can handle occlusions or disappearing of objects during burn-in frames by attending to other timesteps where the objects are visible, and using that information to infer the properties and motion of objects.
}
Finally, the red cylinder merely looks at itself because it is not involved in the collisions.

Figure~\ref{appendix-fig:attn-map} (right) illustrates one example from CLEVRER.
We only analyze the left side of the images since there is no object interaction in the right part.
There are two collision events (the purple cylinder hitting the orange cylinder, and the yellow sphere hitting the blue cube) happening in $\bm{x}_7$, and \algoNameFull successfully captures their interactions in the attention maps.
In general, we found the attention maps in CLEVRER less clear than those in OBJ3D, due to the smaller object size.
Nevertheless, the Transformer can still detect correct cues to reason their future motion.

\begin{figure*}[t]
    \vspace{\pagetopmargin}
    \centering
    \subfigure{
        \includegraphics[width=0.325\textwidth]{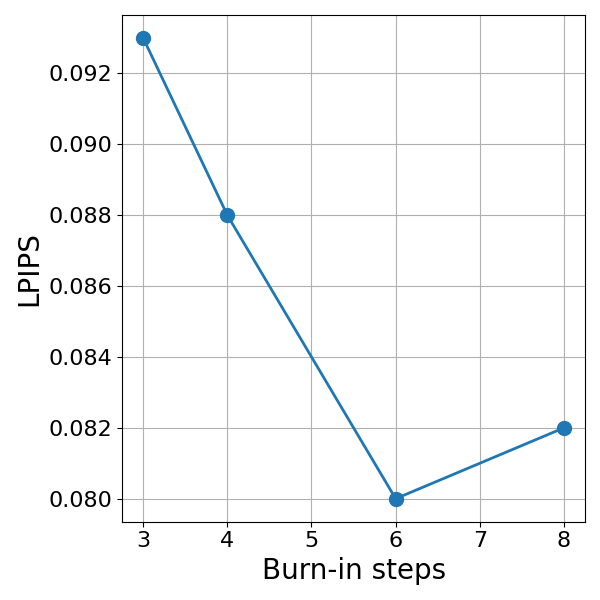}
    }
    \hspace{-3mm}
    \subfigure{
        \includegraphics[width=0.325\textwidth]{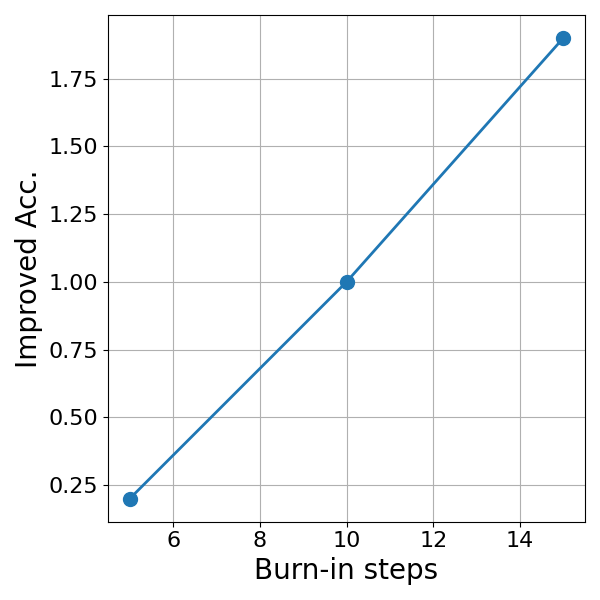}
    }
    \hspace{-3mm}
    \subfigure{
        \includegraphics[width=0.325\textwidth]{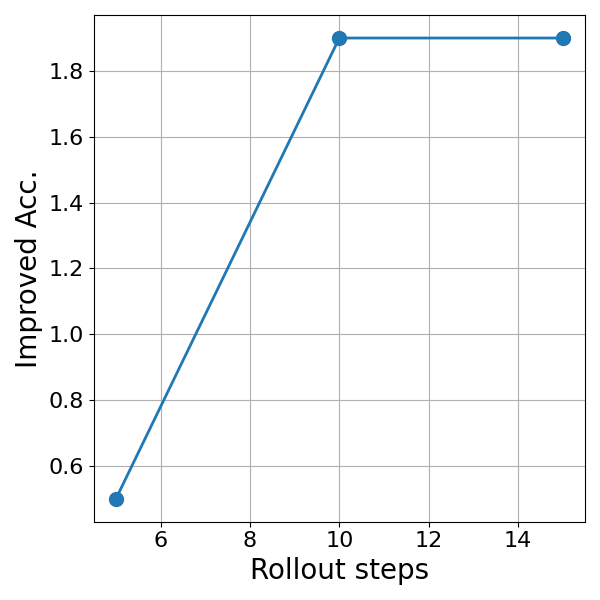}
    }
    \caption{
    \revised{
    Ablation study on burn-in and rollout length of \algoNameFull.
    We show the LPIPS of generated videos on OBJ3D (left), and the improved VQA accuracy by rollout on Physion (middle, right).
    }
    }
    \label{appendix-fig:ablation}
    \vspace{\figmargin}
\end{figure*}

\subsection{\revised{Ablation Study}}\label{appendix:ablation}

\revised{
Figure~\ref{appendix-fig:ablation} shows the effect of burn-in and rollout length on \algoNameFull's performance as line plot for better clarity.
On OBJ3D, the LPIPS first improves as we use more burn-in frames, and then degrades after reaching a peak at $T = 6$.
We do not ablate the rollout length as it is fixed according to the evaluation setting.
On Physion, the accuracy gain increases consistently with more burn-in frames as it provides more context information.
Therefore, we choose the maximum length $T = 15$ according to the number of observed frames available at test time.
Finally, the accuracy grows as we use more rollout frames during training, and plateaus after $K = 10$.
}

\begin{figure*}[t]
    \centering
    \includegraphics[width=0.9\linewidth]{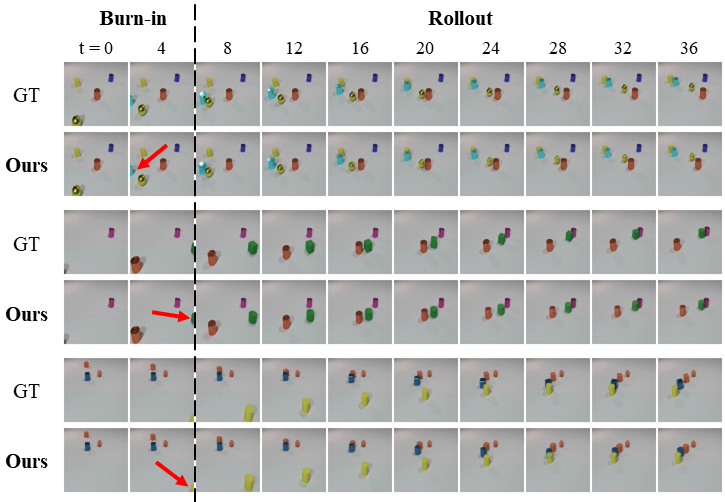}
    \caption{
    \revised{
    Videos generated by \algoNameFull on CLEVRER where some objects are not visible at the initial frame, but enters the scene during burn-in frames (marked by red arrows).
    SAVi is able to detect these new objects, and \algoNameFull can still simulate accurate future dynamics for these objects.
    }
    }
    \label{appendix-fig:burnin-new-obj}
\end{figure*}

\subsection{\revised{New Objects During Burn-in Frames}}

\revised{
One concern regarding \algoNameFull is that, if some objects are not visible at the first timestep, but enters the scene during burn-in frames, can our model still be able to simulate their dynamics?
Figure~\ref{appendix-fig:burnin-new-obj} shows a few examples with new objects appearing during burn-in steps on the CLEVRER dataset.
Since the object-centric model SAVi is able to detect these new objects, \algoNameFull can still reason their dynamics based on frames where they are visible, and simulate accurate future states.
This is an important property of \algoNameFull as disappearing and (re-)appearing of objects are common in real world videos.
}

\section{Limitations \revised{and Future Works}}\label{appendix:limitations}

\heading{Limitations.}
\algoNameFull currently builds upon pretrained object-centric models.
This family of methods still fail to scale up to real world data \footnote{
STEVE~\citep{STEVE} can work on real world videos such as traffics, where ground usually shares the same color, and looks distinct from the vehicles.
Also, STEVE's Transformer-based slot decoder cannot generate images of high visual quality on complex datasets.
}, preventing our application to real world videos as well.
Besides, the two-stage training strategy harms the model performance at the early rollout steps as shown in Figure~\ref{fig:per-step-results}.
It is interesting to explore joint training of the base object-centric model and the Transformer dynamics module, which could potentially benefit the performance of both models.
\revised{
Finally, current \algoNameFull model works in a deterministic manner, and thus cannot model the uncertainty of future dynamics, which is common in real world videos.
}

\revised{
\heading{Future Works.}
We only experiment on unconditional future prediction in this paper.
In the future, we plan to extend \algoNameFull to conditional generation tasks, such as action-conditioned generation as done in~\citet{PARTS}.
Recent works have shown success in this direction by converting conditional inputs to tokens and feeding them to the Transformer~\citep{lookoutside,MotionClip}.
Another direction is to simplify the training process by learning scene decomposition and temporal dynamics jointly.
This may allow the object-centric model to leverage long-term motion cues for unsupervised object discovery~\citep{MotionGroupingVOS}.
Finally, it is important to enable \algoNameFull to learn the multi-modality of future dynamics for stochastic video prediction.
This is key to modeling real world videos faithfully~\citep{VPReview}.
}

\end{document}